\documentclass[runningheads]{llncs}

\usepackage[mobile]{eccv}

\usepackage{eccvabbrv}

\usepackage{graphicx}
\usepackage{booktabs}

\usepackage[accsupp]{axessibility}  % Improves PDF readability for those with disabilities.

\usepackage{hyperref}

\usepackage{orcidlink}
\usepackage{xr}

\usepackage{hyperref}
\usepackage{adjustbox}
\usepackage{graphicx}
\usepackage{caption}
\usepackage{subcaption}
\usepackage{printlen}
\usepackage{float}
\usepackage{wrapfig}
\usepackage{multirow}
\usepackage{booktabs}
\usepackage{pifont}
\usepackage{makecell}
\usepackage{nicefrac}
\usepackage{url}
\usepackage{yhmath}
\usepackage{stmaryrd}
\usepackage{verbatim}
\usepackage{enumitem}
\usepackage{amsmath}
\usepackage{lipsum}
\usepackage{algorithm}
\usepackage{algpseudocode}
\usepackage{amsfonts}
\usepackage{microtype}
\usepackage{pgf}
\usepackage{pgfplots}
\usepackage{tabularx}
\usepackage{scalerel}
\usepackage{xspace}
\usepackage{ifthen}
\usepackage{soul}
\usepackage{cuted}
\usepackage{mathtools}
\usepackage{tikz}
\usetikzlibrary{angles,quotes,3d,math,arrows.meta,calc,positioning,fit,backgrounds,decorations.pathreplacing,calligraphy,shapes,shapes.multipart}
\usepackage[table]{xcolor}
\usepackage{lineno}
\usepackage[makeroom]{cancel}
\usepackage{multirow}
\usepackage{booktabs}
\usepackage{array}
\usepackage{rotating}
\usepackage[font=small,labelfont=bf]{caption}
\RenewDocumentCommand{\paragraph}{s m}{\vspace{1.25em}\noindent\textbf{#2\IfBooleanF{#1}{.}}}

\newcommand{\methodname}{Logit Refiner\xspace}

\newcommand{\cmark}{\ding{51}}
\newcommand{\xmark}{\ding{55}}

\makeatletter
\newcommand{\xdashrightarrow}[2][]{\ext@arrow 0359\rightarrowfill@@{#1}{#2}}
\newcommand{\xdashleftarrow}[2][]{\ext@arrow 3095\leftarrowfill@@{#1}{#2}}
\newcommand{\xdashleftrightarrow}[2][]{\ext@arrow 3359\leftrightarrowfill@@{#1}{#2}}
\def\rightarrowfill@@{\arrowfill@@\relax\relbar\rightarrow}
\def\leftarrowfill@@{\arrowfill@@\leftarrow\relbar\relax}
\def\leftrightarrowfill@@{\arrowfill@@\leftarrow\relbar\rightarrow}
\def\arrowfill@@#1#2#3#4{%
  $\m@th\thickmuskip0mu\medmuskip\thickmuskip\thinmuskip\thickmuskip
   \relax#4#1
   \xleaders\hbox{$#4#2$}\hfill
   #3$%
}
\makeatother

\definecolor{ourgreen}{RGB}{46, 204, 113}
\definecolor{ourgreenborder}{RGB}{39, 174, 96}
\definecolor{ourblue}{RGB}{52, 152, 219}
\definecolor{ourblueborder}{RGB}{41, 128, 185}
\definecolor{ourorange}{RGB}{230, 126, 34}
\definecolor{ourorangeborder}{RGB}{211, 84, 0}
\definecolor{ourred}{RGB}{231, 76, 60}
\definecolor{ourredborder}{RGB}{192, 57, 43}
\definecolor{ouryellow}{RGB}{241, 196, 15}
\definecolor{ouryellowborder}{RGB}{243, 156, 18}
\definecolor{ourpurple}{RGB}{155, 89, 182}
\definecolor{ourpurpleborder}{RGB}{142, 68, 173}
\definecolor{ourturquoise}{RGB}{26, 188, 156}
\definecolor{ourturquoiseborder}{RGB}{22, 160, 133}
\definecolor{ourturquoise}{RGB}{26, 188, 156}
\definecolor{ourturquoiseborder}{RGB}{22, 160, 133}
\definecolor{ourwhite}{RGB}{236, 240, 241}
\definecolor{ourwhiteborder}{RGB}{189, 195, 199}
\definecolor{ourgray}{RGB}{149, 165, 166}
\definecolor{ourgrayborder}{RGB}{127, 140, 141}

\definecolor{ourwhite2}{RGB}{246, 247, 248}

\definecolor{matplotlibblue}{HTML}{1f77b4}
\definecolor{matplotliborange}{HTML}{ff7f0e}
\definecolor{matplotlibgreen}{HTML}{2ca02c}

\definecolor{ourhighlightcolor}{RGB}{46, 204, 113}

\newcolumntype{H}{>{\setbox0=\hbox\bgroup}c<{\egroup}@{}}

\newcommand{\tikzstylenodedistance}{4mm}
\newcommand{\tikzstyleinnersep}{2mm}
\newcommand{\tikzstyleminimumheight}{8.75mm}
\newcommand{\tikzstyleminimumwidth}{12mm}

\tikzset{
    node distance=\tikzstylenodedistance,
    text centered,
    anchor=center,
}
\tikzset{
    standard node/.style n args={1}{%
        rectangle,
        rounded corners=0.1cm,
        fill=our#1,
        draw=our#1border,
        line width=0.04cm,
        minimum height=\tikzstyleminimumheight,
        minimum width=\tikzstyleminimumwidth,
        inner sep=\tikzstyleinnersep,
        text centered,
        anchor=center,
        align=center,
    }
}
\tikzset{
    standard node module/.style n args={0}{%
        rectangle,
        rounded corners=0.1cm,
        fill=ourturquoise,
        draw=ourturquoiseborder,
        line width=0.04cm,
        minimum height=\tikzstyleminimumheight, % 10.5
        minimum width=12mm, %\tikzstyleminimumwidth, % 16
        inner xsep=\tikzstyleinnersep,
        inner ysep=1mm,
        text centered,
        anchor=center,
        align=center,
    }
}
\tikzset{
    standard node image/.style n args={1}{%
        rectangle,
        fill=our#1,
        draw=our#1border,
        line width=0.04cm,
        minimum height=\tikzstyleminimumheight,
        minimum width=\tikzstyleminimumwidth,
        inner sep=0,
        text centered,
        anchor=center,
        align=center,
    }
}
\tikzset{
    standard node circle/.style n args={1}{%
        fill=our#1,
        draw=our#1border,
        circle,
        inner sep=0.1cm,
        minimum height=0,
        minimum width=0,
    }
}
\tikzset{
    standard node circle/.prefix style = standard node
}

\tikzset{
    standard line/.style n args={0}{%
        line width=0.04cm,
        rounded corners=0.1cm,
    }
}
\tikzset{
    standard arrow/.style n args={0}{%
        -latex,
    }
}
\tikzset{
    standard arrow/.prefix style = standard line
}

\tikzset{
    simple node image/.style n args={0}{%
        rectangle,
        inner sep=0,
        text centered,
        anchor=center,
        align=center,
        node distance=0mm
    }
}

\usepackage{xparse}
\usepackage[capitalize]{cleveref}

\RenewDocumentCommand{\paragraph}{s m}{\vspace{1.25em}\noindent\textbf{#2\IfBooleanF{#1}{.}}}

\makeatletter
\ExplSyntaxOn
\NewDocumentCommand{\citep}{ o o m }
  {
    \begingroup
    \def\@cite##1##2{%
      [%
        \IfNoValueTF{#1}{}{ \tl_if_blank:nF {#1} {#1\nobreakspace} }%
        ##1%
        \tl_if_blank:nF {##2} {,\nobreakspace ##2}%
      ]%
    }%
    \IfNoValueTF{#2}
      { \cite{#3} }
      { \tl_if_blank:nTF {#2} { \cite{#3} } { \cite[#2]{#3} } }
    \endgroup
  }
\ExplSyntaxOff
\makeatother

\makeatletter
\newcommand{\citet@saveauthor}[2]{%
  \expandafter\gdef\csname citet@author@#1\endcsname{#2}%
}
\AtBeginDocument{%
  \let\citet@orig@@bibitem\@bibitem
  \def\@bibitem#1{%
    \citet@orig@@bibitem{#1}%
    \citet@scanauthor{#1}%
  }%
}
\long\def\citet@scanauthor#1 #2:{%
  \citet@extractname{#1}#2\citet@endmark
  :%
}
\def\citet@extractname#1#2,#3\citet@endmark{%
  \expandafter\gdef\csname citet@author@#1\endcsname{#2}%
  \if@filesw
    \protected@write\@auxout{}{\string\citet@saveauthor{#1}{#2}}%
  \fi
  #2,#3%
}
\DeclareDocumentCommand{\citet}{m}{%
  \ifcsname citet@author@#1\endcsname
    \csname citet@author@#1\endcsname~et~al.~\cite{#1}%
  \else
    \textbf{??}~\cite{#1}%
  \fi
}
\makeatother

\makeatletter
\renewcommand\subsubsection{%
  \@startsection{subsubsection}{3}{\z@}%
    {-12pt \@plus -4pt \@minus -2pt}%
    {4pt \@plus 2pt \@minus 2pt}%
    {\normalfont\normalsize\bfseries}%
}
\makeatother

\newcommand{\samebf}[1]{\begingroup\sbox0{#1}\sbox2{\textbf{#1}}\resizebox{\wd0}{\ht2}{\usebox2}\endgroup}

\crefname{appsec}{Supp.\ Sec.}{Supp.\ Secs.}
\Crefname{appsec}{Supp.\ Section}{Supp.\ Sections}
\crefname{appfig}{Supp.\ Fig.}{Supp.\ Figs.}
\Crefname{appfig}{Supp.\ Figure}{Supp.\ Figures}
\crefname{apptab}{Supp.\ Tab.}{Supp.\ Tabs.}
\Crefname{apptab}{Supp.\ Table}{Supp.\ Tables}
\crefname{appeq}{Supp.\ Eq.}{Supp.\ Eqs.}
\Crefname{appeq}{Supp.\ Equation}{Supp.\ Equations}

\begin{document}

% ---------------------------------------------------------------
% TODO REVIEW: Replace with your title
% \title{\methodname: Improving Discrete Image Generation with \todo{Chunkwise} Dependency Modeling} 
\title{\methodname: {Improving Visual Autoregressive Models via Intra-Scale Dependency Modeling}}

% TODO REVIEW: If the paper title is too long for the running head, you can set
% an abbreviated paper title here. If not, comment out.
\titlerunning{{\methodname}}

% % TODO FINAL: Replace with your author list. 
% % Include the authors' OCRID for the camera-ready version, if at all possible.
% \author{First Author\inst{1}\orcidlink{0000-1111-2222-3333} \and
% Second Author\inst{2,3}\orcidlink{1111-2222-3333-4444} \and
% Third Author\inst{3}\orcidlink{2222--3333-4444-5555}}

% % TODO FINAL: Replace with an abbreviated list of authors.
% \authorrunning{F.~Author et al.}
% % First names are abbreviated in the running head.
% % If there are more than two authors, 'et al.' is used.

% % TODO FINAL: Replace with your institution list.
% \institute{Princeton University, Princeton NJ 08544, USA \and
% Springer Heidelberg, Tiergartenstr.~17, 69121 Heidelberg, Germany
% \email{lncs@springer.com}\\
% \url{http://www.springer.com/gp/computer-science/lncs} \and
% ABC Institute, Rupert-Karls-University Heidelberg, Heidelberg, Germany\\
% \email{\{abc,lncs\}@uni-heidelberg.de}}
\author{
    Meimingwei Li\thanks{Equal contribution.}\inst{,1}
    \and Stefan Andreas Baumann\textsuperscript{$\star$}\inst{,1,2}
    \and \\Felix Krause\inst{1,2}
    \and Bj\"orn Ommer\inst{1,2}
}
\authorrunning{Li \& Baumann et al.}
\institute{CompVis @ LMU Munich, Germany \and Munich Center for Machine Learning (MCML)}

\maketitle

\begin{abstract}
    Visual Autoregressive Models (VAR) generate images through next-scale prediction, producing all tokens within each scale in parallel.
We show that this parallel decoding constitutes a mean-field-style approximation that discards spatial dependencies among same-scale tokens, causing locally incoherent samples regardless of backbone capacity -- a limitation of the \emph{decoding rule}.
Addressing this limitation, we introduce the \emph{\methodname}, a lightweight autoregressive module that restores intra-scale dependencies by sequentially sampling tokens conditioned on frozen backbone features.
Adding only $\sim$10\% parameters and less than 5\% of the base model's training compute, it plugs into any pretrained VAR checkpoint without retraining.
Controlled ablations isolate joint intra-scale sampling -- rather than additional capacity or training -- as the critical ingredient.
Across backbones from 310M to 2B parameters on class-conditional ImageNet $256{\times}256$, the refiner consistently improves generation quality, enabling a 1.1B-parameter model to surpass one twice its size.
The approach further generalizes to text-to-image generation, confirming that the mean-field bottleneck persists across VAR variants and is effectively alleviated by our method.\\
Project page: \href{https://compvis.github.io/logit-refiner/}{https://compvis.github.io/logit-refiner/}.
    % % \keywords{Image Generation \and Scale-wise Autoregression \and Visual Autoregressive Modeling}
    % \keywords{Generative Models \and Image Generation \and Scale-wise Autoregression}
\end{abstract}

\vspace{-1em}

\begin{figure}[t]
    \centering
    \includegraphics[width=\linewidth]{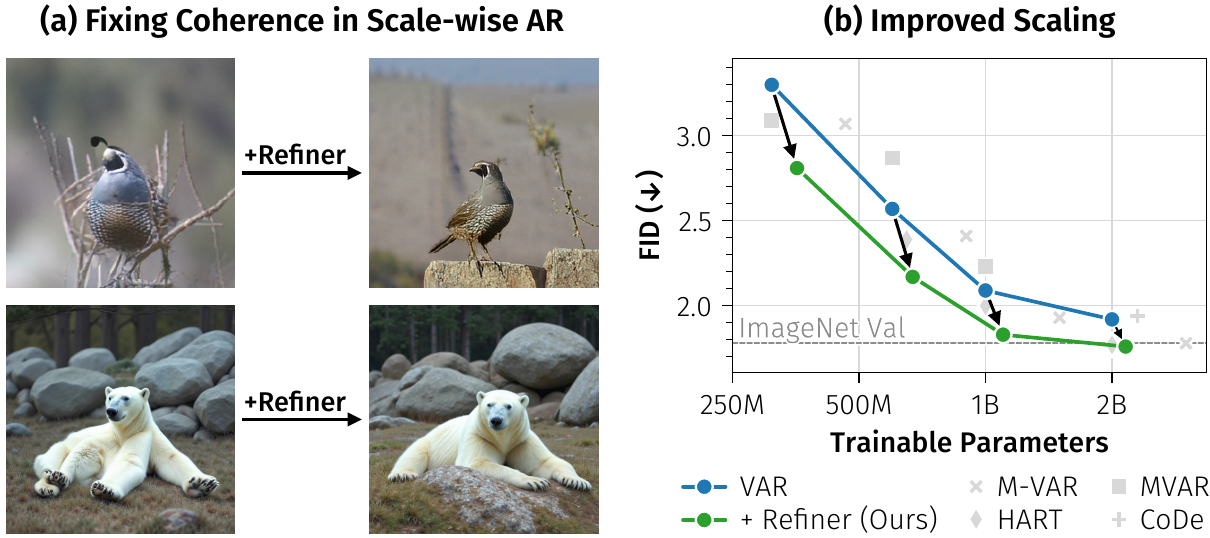}
    \caption{\textbf{(a)} VAR~\cite{tian2024visual} (top) and scaled-up variants like Infinity~\cite{han2025infinity} (bottom) often generate spatially incoherent samples. Our \methodname addresses this problem via a lightweight add-on to existing pretrained models. \textbf{(b)} The refiner consistently improves scaling behavior, shifting the VAR scaling curve downward, until saturating at the same FID that true unseen samples (validation set) achieve.}
    \label{fig:teaser}
\end{figure}

\section{Introduction}
Autoregressive (AR) modeling achieves remarkable success in language~\cite{brown2020language,achiam2023gpt,touvron2023llama,team2023gemini,bai2023qwen,bi2024deepseek,sun2021ernie,chowdhery2023palm,anil2023palm,hoffmann2022training} by generating tokens sequentially from a learned joint distribution.
Extending this paradigm to images, however, is challenging: images lack a canonical ordering~\cite{van2016conditional,salimans2017pixelcnn++,parmar2018imagetrasnformer,wang2025gpt}, and fully sequential pixel generation~\cite{van2016conditional,salimans2017pixelcnn++,parmar2018imagetrasnformer,wang2025gpt} is prohibitively slow, as sequence lengths quickly reach tens of thousands of tokens.

Visual Autoregressive Modeling (VAR)~\cite{tian2024visual} resolves this by leveraging \emph{next-scale prediction}, generating images in a coarse-to-fine manner, while predicting all tokens within each scale in parallel.
This design yields strong ~\cite{russakovsky2015imagenet} generation results,
scaling behavior, and substantially improved efficiency compared to token-wise autoregressive models.

Despite accurate per-token predictions, VAR samples often exhibit \emph{local spatial incoherence}:
neighboring patches display mismatched textures, structural discontinuities, or implausible combinations -- even when each individual prediction is plausible (\cref{fig:var_failures}).
This raises a fundamental question:
\begin{center}
    \emph{If the per-token marginals are correct, why are the joint samples incoherent?}
\end{center}
We identify the root cause as an implicit \emph{mean-field-style assumption} in VAR's parallel within-scale decoding, which factorizes the conditional joint into independent per-token distributions, discarding spatial dependencies among tokens at the same scale.
This approximation produces token combinations that are individually likely, yet jointly inconsistent.
A minimal checkerboard example~(\cref{fig:checkerboard}) makes this concrete:
correct per-pixel probabilities still yield invalid global patterns under independent sampling.

Crucially, this limitation lies in the \emph{decoding rule} -- even predicting the correct pointwise conditional distributions can not yield valid samples in practice, since tokens are decoded independently.
The missing component in scale-wise autoregressive image generation is therefore \emph{joint within-scale sampling} that accounts for the intra-scale token dependencies.
We introduce the \emph{\methodname}, a lightweight \emph{add-on} autoregressive module that restores intra-scale dependencies while leaving the VAR backbone untouched.
It can directly be applied on top of an already pretrained VAR model, without any need for adaptation of the pretrained weights to obtain the quality benefits.
Conditioned on the backbone's already-computed hidden states, the refiner samples tokens sequentially within each scale, requiring only a small causal model to capture the residual dependencies that independent decoding ignores.
The module adds only $\sim$10\% parameters, trains in hours ($<$5\% extra training compute) with the backbone frozen, and plugs into any pretrained VAR model with modest inference overhead.

Across backbone sizes from 310M to 2B parameters on class-conditional ImageNet, the \methodname consistently improves generative performance by a wide margin, enabling VAR-d24 + Refiner (1.1B parameters) to surpass the twice as large VAR-d30 (2B).
Controlled ablations confirm that these gains stem from \emph{dependency modeling rather than additional capacity or training}: an architecture-matched refiner with bidirectional attention and independent sampling fails to match the autoregressive variant, isolating joint intra-scale sampling as the critical ingredient.
The approach also generalizes to text-to-image generation, where the refiner yields consistent improvements on a 2B-parameter model.

Our work makes the following main contributions:
starting by tracing the spatial incoherence observed in VAR samples to a specific cause -- the mean-field-style approximation inherent in parallel within-scale decoding, which generates tokens independently regardless of the backbone's capacity --
we show that autoregressive within-scale sampling is the minimal correction needed to remove this approximation error, reframing the problem from model capacity to the decoding rule.
Then, we introduce a \emph{lightweight} autoregressive refiner that implements this correction as a plug-in module over frozen backbone features, consistently improving generation quality across model scales from 310M to 2B parameters.
\section{Related Work}

\paragraph{Autoregressive Image Generation}
Autoregressive models are a dominant paradigm for image generation, powering many frontier foundation models~\cite{google2025nanobanana,google2025nanobananapro,openai2025gptimage1,openaigptimage15,xai2025grokimagine}.
Early approaches modeled images as pixel-level sequences~\cite{theis2015generative,oord2016pixel,van2016conditional,salimans2017pixelcnn++}, whereas modern methods~\cite{chen2020imagegpt,ramesh2021zero,yu2022scaling,sun2024autoregressive,lee2022autoregressive,tian2024visual,yu2021vector} operate on discrete tokens from learned tokenizers~\cite{van2017neural,razavi2019generating,esser2021taming,yu2023language} and explore alternatives to the standard ``sweep'' ordering~\cite{esser2021taming,yu2025randomized,xu2025direction}, the grouping of multiple tokens~\cite{tian2024visual,chang2022maskgit,wang2025parallelized,ren2025beyond}, or shared backbones with large language models~\cite{team2024chameleon,sun2024emu,sun2024generative,wang2026multimodal,wu2025janus,chen2025janus}.
Our work is orthogonal: rather than changing the token order or the backbone, we correct the independence assumption within parallel decoding groups.

\paragraph{Scale-wise Autoregressive Image Generation}
VAR~\cite{tian2024visual} introduced next-scale prediction, generating tokens at progressively finer resolutions while sampling all tokens within each scale in parallel. The paradigm has since been extended in many directions: Infinity and Switti scale it to text-to-image generation~\cite{han2025infinity,liu2025infinitystar,voronov2024switti}; M-VAR~\cite{ren2024m} and MVAR~\cite{zhang2026mvar} introduce more efficient backbones, and HMAR~\cite{kumbong2025hmar} combines next-scale prediction with masked autoregressive modeling~\cite{chang2022maskgit}; FVAR~\cite{li2025fvar}, HART~\cite{tang2024hart}, and FlowAR~\cite{ren2024flowar} alter the prediction target or token representation; and a growing line of work reduces inference cost through token pruning~\cite{guo2025fastvar,chen2025frequency,li2026sparvar}, frequency- or entropy-guided skipping~\cite{lifreqexit,chen2026toprovar,li2025skipvar,zhang2026adaptive}, and speculative decoding~\cite{chen2025collaborative}.
Beyond class-conditional generation, the paradigm has been applied to multimodal modeling~\cite{zhuang2025vargpt,zhuang2025vargpt1}, image editing~\cite{mao2025visual,dao2025discrete}, restoration~\cite{rajagopalan2025restorevar}, super-resolution~\cite{qu2025visual}, segmentation~\cite{zheng2025seg}, and video generation~\cite{ji2026videoar,liu2025infinitystar}.
Across these extensions, tokens within each scale remain at least partially decoded independently, suggesting that the mean-field-style approximation is a structural property of the scale-wise autoregressive paradigm rather than a task-specific limitation of VAR.

\paragraph{Refinement and Post-hoc Correction}
Refining initial predictions appears in many forms, from draft-and-revise generation that iteratively improves masked tokens~\cite{lee2022draft,zheng2025lsrs} to image generators with explicit refinement stages that re-predict a full generated image~\cite{podell2024sdxl,wang2025visual}.
Other methods~\citep[cf.][]{bi2026adversarialerrorcorrectionvisual} propose adversarially optimizing guidance injection into the decoding rule to improve sample quality.
Unlike these, we target a specific shortcoming of VAR-style models~\cite{tian2024visual} -- the mean-field-style approximation -- and correct it with a lightweight add-on inside the generation loop, avoiding repeated full-model passes over the entire image.

\section{Mean-Field-style Approximation in Scale Autoregression}
\label{sec:method_var_prelim}
Visual Autoregressive Modeling (VAR)~\cite{tian2024visual} generates images through \emph{next-scale prediction} over a hierarchy of discrete token maps.
An image is encoded into $K$ scales $\mathbf{r}_{1:K} = (\mathbf{r}_1, \ldots, \mathbf{r}_K)$, where each $\mathbf{r}_k \in [V]^{L_k}$ is a sequence of $L_k = h_kw_k$ tokens at spatial resolution $h_k \times w_k$.
VAR models the joint distribution autoregressively \emph{across scales}:
\begin{equation}
    \label{eq:var_scale_factorization}
    p(\mathbf{r}_{1:K}) = \prod_{k=1}^K p_\theta(\mathbf{r}_k \mid \mathbf{r}_{<k}).
\end{equation}
At each scale, a transformer backbone $f_\theta$ produces hidden states $\mathbf{h}^{(k)} = f_\theta(\mathbf{r}_{<k})$ for all token positions in parallel that parametrize per-token categorical distributions.
Tokens within each scale are then sampled \emph{independently}:
\begin{equation}
    \label{eq:var_mean_field_factorization}
    p_\theta(\mathbf{r}_k \mid \mathbf{r}_{<k}) \approx \prod_{i=1}^{L_k} p_\theta\!\left(r_i^{(k)} \mid \mathbf{h}^{(k)}\right),
\end{equation}
which constitutes a fully factorized (naive) \emph{mean-field-style approximation} that ignores spatial dependencies across tokens in the same scale, akin to those typically used in variational inference in physics~\cite{jordan1999introduction}.
Natural images, however, exhibit strong spatial dependencies, implying that the true conditional joint $p_\theta(\mathbf{r}_k \mid \mathbf{r}_{<k})$ might \emph{not} be modeled faithfully.
This can lead to structurally incoherent samples that persist even at the largest model scales (\cref{fig:var_failures}).

\begin{figure}[t]
    \centering
    \includegraphics[width=\linewidth]{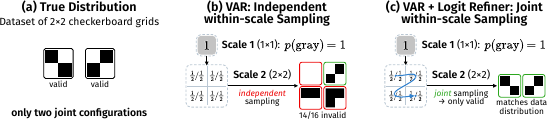}
    \caption{\textbf{Toy Example.} \textbf{(a)} Consider a dataset of $2\times 2$ checkerboards. \textbf{(b)}~A VAR-style model can learn correct per-token marginals (gray at $1^2$, 50/50 at $2^2$), yet independent sampling yields many invalid joint samples. \textbf{(c)} Our Logit Refiner models remaining dependencies autoregressively, restoring the joint and generating only valid samples.}
    \label{fig:checkerboard}
\end{figure}

\paragraph{Toy Example (\cref{fig:checkerboard})}
For a simple dataset containing only valid $2\times 2$ checkerboards (two valid samples), a next-scale model predicts correct per-token marginals, yet independent sampling produces $2^4 = 16$ joint outcomes, most of which are invalid.
This illustrates the central failure of \cref{eq:var_mean_field_factorization}: \emph{correct marginals do not imply correct joint samples}, even when conditioning on previous scales.

This limitation arises from the \emph{decoding rule}, motivating the restoration of the intra-scale joint distribution.

\section{\methodname}
We introduce the \emph{\methodname}, a lightweight autoregressive module that restores intra-scale dependencies while leaving the pretrained VAR backbone unchanged.
The backbone $f_\theta$ continues to compute per-token hidden states in parallel, whereas a separate refiner model $q_\phi$ models the joint distribution of tokens within each scale conditioned on these frozen features.
Since the refiner operates purely at the decoding stage and is not constrained by VAR's mean-field-style factorization, it can implement flexible joint sampling with minimal additional parameters, computation, and training cost.

\subsection{Restoring Joint Within-Scale Sampling}
Since the backbone's bidirectional intra-scale attention already enables joint reasoning about all token positions per scale, the independence limitation resides in the \emph{sampling rule}, not the learned representations (\cref{sec:method_var_prelim}).
We therefore replace the mean-field-style decoder in \cref{eq:var_mean_field_factorization} with an autoregressive factorization over tokens within each scale:
\begin{equation}
    \label{eq:refiner_ar_factorization}
    q_\phi(\mathbf{r}_k \mid \mathbf{r}_{<k}) = \prod_{i=1}^{L_k} q_\phi\!\left(r_i^{(k)} \mid r^{(k)}_{<i},\, \mathbf{h}^{(k)}_{\leq i},\, \mathbf{r}_{<k}\right),
\end{equation}
where $q_\phi$ is a lightweight \emph{refiner} model.
Each token now conditions on both the frozen backbone features $\mathbf{h}^{(k)}_{\leq i}$ and the previously sampled tokens $r^{(k)}_{<i}$, restoring the intra-scale joint that the parallel mean-field-style decoder discards.
This formulation strictly generalizes the original decoder -- setting the refiner in \cref{eq:refiner_ar_factorization} to ignore the autoregressive context directly recovers \cref{eq:var_mean_field_factorization} -- making it a \emph{minimal correction} that removes the conditional independence assumption without modifying the backbone.

Combining \cref{eq:refiner_ar_factorization} with the scale-wise factorization \cref{eq:var_scale_factorization} yields the full model:
\begin{equation}
    p(\mathbf{r}_{1:K}) \approx \prod_{k=1}^K q_\phi(\mathbf{r}_k \mid \mathbf{r}_{<k}),
\end{equation}
which preserves VAR's efficient across-scale generation while restoring within-scale dependencies.
Unlike a fully autoregressive model that runs the entire backbone per token, only the lightweight refiner $q_\phi$ operates sequentially on a token level -- the expensive backbone computation remains fully parallel.
Because the refiner operates on the backbone's features rather than building context from scratch, it only needs to model \emph{residual} dependencies, explaining why an extremely small model suffices (\cref{sec:method_architecture_blocks}).

\begin{figure}[t]
    \centering
    \includegraphics[width=\linewidth]{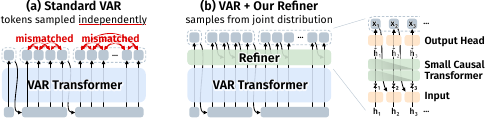}
    \caption{\textbf{Logit Refiner Overview.} \textbf{(a)} The VAR backbone processes all previous scales and produces hidden states for the current scale in a single parallel forward pass, finally sampling from pointwise posteriors in parallel. As sampling is done independently within each scale, this can lead to mismatched tokens, affecting generation quality. \textbf{(b)} Our logit refiner takes these hidden states and samples tokens autoregressively within the scale, conditioning each prediction on previously sampled tokens. The refiner is a lightweight causal transformer, incurring only a small overhead during generation, while significantly improving sample quality.}
    \label{fig:refiner_overview}
\end{figure}

\subsection{Architecture}
The refiner is designed as a strict add-on: it never re-encodes the image and only consumes i) the already-computed backbone hidden states $\mathbf{h}^{(k)}$ for the current scale, and ii) the previously generated tokens \emph{within} the current chunk $r^{(k)}_{<i}$.
This separation ensures that the expensive backbone forward pass remains fully parallel; only the lightweight refiner runs sequentially within each scale.
We describe each component below and illustrate the architecture in \cref{fig:refiner_overview}.

\paragraph{Inputs}
For each position $i$ in scale $k$, the refiner constructs an input vector from two streams of information: the frozen backbone hidden state $\mathbf{h}_i^{(k)} \in \mathbb{R}^w$, and an autoregressive context embedding $\mathbf{c}_i^{(k)} \in \mathbb{R}^w$ derived from the previously sampled token:
\begin{equation}
    \label{eq:refiner_input}
    \mathbf{c}_i^{(k)} = \begin{cases}
        % \mathbf{z}_{[\mathrm{SOS}]}
        \,\mathbf{z}_\mathrm{sos}, & i = 1, \\
        \,\mathrm{emb}_\phi(r_{i-1}^{(k)}), &i > 1,
    \end{cases}
    \qquad
    \mathbf{z}_i^{(k)} = \mathbf{W}_\mathrm{proj}[\,\mathbf{h}_i^{(k)} \parallel \mathbf{c}_i^{(k)}\,],
\end{equation}
with learned $\mathbf{z}_\mathrm{sos}$, token embedding $\mathrm{emb}_\phi : [V] \rightarrow \mathbb{R}^w$, and input projection $\mathbf{W}_\mathrm{proj}$, and $[\;\cdot\parallel\cdot\;]$ denoting concatenation.

\paragraph{Refiner Blocks}\label{sec:method_architecture_blocks}
A small stack of $d_r$ transformer blocks processes the fused representations $\mathbf{z}_{1:L_k}^{(k)}$ with a causal mask, producing refined hidden states
\begin{equation}
    \label{eq:refiner_blocks}
    \tilde{\mathbf{h}}_i^{(k)} = \mathrm{TransformerBlocks}_\phi\bigl(\mathbf{z}_{1:i}^{(k)}\bigr).
\end{equation}
Each block follows standard transformer design, matching the backbone's block architecture.
Crucially, only $d_r \ll d$ blocks are needed (e.g., $d_r = 2$) -- far fewer than the backbone depth ($d \in [16, 30]$), since the refiner already receives a rich, spatially-contextualized hidden state $\mathbf{h}_i^{(k)}$ from the backbone with bidirectional attention.
The refiner only needs to model the \emph{residual} dependencies not captured by the backbone -- a much easier task than building spatial context from scratch.

The autoregressive factorization in \cref{eq:refiner_ar_factorization} requires choosing a token ordering within each scale; we use standard raster-scan order (left-to-right, top-to-bottom), following conventions in patch-level autoregressive models~\citep[cf.][]{oord2016pixel,van2016conditional}.

\paragraph{Output}
An output head predicts logits from the refined hidden state, from which a token is sampled:
\begin{equation}
    \label{eq:refiner_output}
    \tilde{\ell}_i^{(k)} = \mathrm{head}_\phi\bigl(\tilde{\mathbf{h}}_i^{(k)}\bigr),
    \qquad
    r_i^{(k)} \sim \mathrm{Cat}\bigl(\mathrm{softmax}\bigl(\tilde{\ell}_i^{(k)}\bigr)\bigr).
\end{equation}
The sampled token $r_i^{(k)}$ is then fed back as the autoregressive context for the next position via $\mathrm{emb}_\phi$ in \cref{eq:refiner_input}, and this process repeats sequentially for all $L_k$ positions in the scale.

\subsection{Training and Inference}
Starting from a conventionally pretrained VAR model, we train the refiner with teacher forcing: during training, the autoregressive context $r^{(k)}_{<i}$ in \cref{eq:refiner_input} is replaced with ground-truth tokens.
Combined with the causal attention mask, this allows training to be fully parallelized across all positions and scales simultaneously.
We minimize cross-entropy over all tokens:
\begin{equation}
    \mathcal{L}(\phi) = -\sum_{k=1}^K \sum_{i=1}^{L_k} \log q_\phi\!\left(r_i^{(k)} \mid r^{(k)}_{<i},\, \mathbf{h}^{(k)}_{\leq i},\, \mathbf{r}_{<k}\right).
\end{equation}
During refiner training, the VAR backbone $f_\theta$ remains \emph{frozen}.
We only optimize the refiner parameters $\phi$, finding that this suffices for achieving significant performance gains while keeping training cost minimal.

\begin{wrapfigure}{r}{.33\linewidth}
    \centering
    \vspace{-2.2em}
    \includegraphics[width=\linewidth]{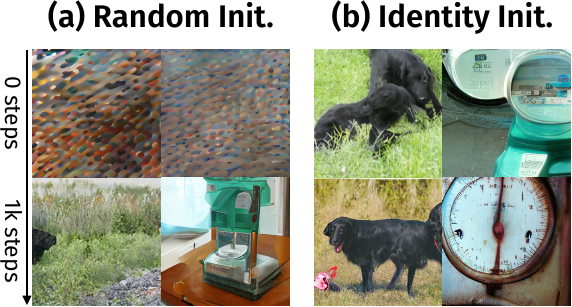}
    \caption{\textbf{Effect of Initialization Strategy.}}
    \label{fig:initialization_strategy}
\end{wrapfigure}

\paragraph{Identity Initialization}
We design an initialization scheme that makes the refiner reproduce the base model's predictions at the start of training, so that optimization focuses entirely on learning the residual corrections needed for joint sampling.
Specifically, we copy the pretrained output head and token embedding weights from the base model, providing the refiner with an already well-structured output space.
The autoregressive context integration via $\mathbf{W}_\mathrm{proj}$ (in \cref{eq:refiner_input}) is initially disabled by setting $\mathbf{W}_\mathrm{proj} \gets [\mathbf{I}\parallel\mathbf{0}]$; similarly, the output projections of each transformer block's self-attention and feedforward networks are zero-initialized to leave the initial hidden states unchanged.
Under this scheme, the model at initialization is functionally equivalent to the original VAR, and the training signal drives the refiner to learn only the \emph{difference} between independent and joint within-scale distributions.
As shown in \cref{fig:initialization_strategy}, identity initialization retains the base model's generation quality from the first iteration and converges significantly faster than standard random initialization.

\paragraph{Inference}
At inference, VAR generates tokens scale-wise in a coarse-to-fine manner.
For each scale, we compute $\mathbf{h}^{(k)}$ once in parallel using the expensive base model $f_\theta$.
Then, we sample tokens $r_1^{(k)}, \ldots, r_{L_k}^{(k)}$ sequentially using KV caching in the refiner.
Since the refiner is much smaller than the base model ($d_r \ll d$), the additional cost is modest and significantly lower than traditional full-model tokenwise autoregressive sampling.

\section{Experiments}

\paragraph{Experiment Settings}
We conduct our experiments on class-conditional ImageNet~\cite{deng2009imagenet} at a resolution of $256^2$ unless noted otherwise, and primarily evaluate the Fr\'echet Inception Distance (FID)~\cite{heusel2017gans} on 50k generated samples.
Our base model implementation and training setting directly follow VAR~\cite{tian2024visual}, except for a significantly reduced training duration of 30 epochs compared to multiple hundred for the base models, and a reduced base learning rate (1e-4 $\rightarrow$ 2.5e-5).
Unless noted otherwise, we use $d_r = 2$ refiner layers whose width is matched to the base model's parameters.
For sampling, we use classifier-free guidance (CFG)~\cite{ho2022classifier} and top-k sampling following the original VAR settings, with individually swept parameters.
Full hyperparameters and details are provided in \cref{sec:app_implementation_details}.

\subsection{Ablation Studies}
We first validate the key design decisions of the \methodname through controlled ablations, establishing that the gains originate from dependency modeling, before presenting final results.

\begin{table}[t]
    \centering
    \caption{\textbf{What drives the gains?} Neither additional training nor additional parallel layers match the improvement from autoregressive (joint) intra-scale refinement. All variants use VAR-d16 as backbone, the ``parallel'' and ``AR'' refiner use the same architecture ($d_r = 2$), with only a different attention mask and sampling. Only causal attention with autoregressive sampling -- i.e., joint intra-scale modeling -- yields substantial gains.}
    \adjustbox{max width=\linewidth}{
    \begin{tabular}{lccc}
        \toprule
        Model & Joint Modeling & Params & FID$\downarrow$ \\
        \midrule
         VAR-d16 (baseline)~\cite{tian2024visual} & \xmark & 310M & 3.30 \\
        + Additional Training (30ep) & \xmark & 310M & 3.12 \\
        + Parallel Refiner (bidirectional attention) & \xmark & 356M & 3.15 \\
        + \textbf{AR Refiner (causal attention, ours)} & \cmark & 356M & 2.81 \\
        \bottomrule
    \end{tabular}
    }
    \label{tab:ablation_gains}
\end{table}

\paragraph{Dependency Modeling, not Capacity, Drives Gains}
\Cref{tab:ablation_gains} disentangles the contribution of joint intra-scale modeling from additional capacity and training on the same pretrained VAR-d16 backbone, exploring the following variations:
\begin{enumerate}
    \item \textbf{Additional Training:} continue training the base model for 30 more epochs without any architectural changes.
    \item \textbf{Parallel Refiner:} add refiner-sized transformer blocks with bidirectional attention to the frozen backbone, matching our method's architecture and parameter count but sampling all tokens independently per scale, isolating capacity from joint modeling.
    \item \textbf{AR Refiner (Ours):} identical architecture with causal attention and autoregressive sampling, restoring intra-scale dependencies.
\end{enumerate}
All three variants are trained for the same number of epochs.
Only the full AR refiner yields substantial gains, confirming that \emph{dependency modeling} -- not capacity or extra training -- is the critical ingredient.

\begin{table}[t]
    \centering
    \caption{\textbf{Refiner Design Ablation.} We start from VAR-d16~\cite{tian2024visual}. \textbf{(a)} Refiner depth: $d_r=0$ (no transformer blocks) already improves FID significantly; $d_r = 2$ saturates quality. \textbf{(b)} Trainable components: the refiner alone captures most gains; jointly training the backbone adds little at much higher cost. \textbf{(c)} Per-scale importance: removing the refiner from any single scale degrades quality, most strongly at early scales; degradations diminish at high resolutions.}
    \label{tab:main_ablations}
    \newlength{\mainabltabheight}
    \setlength{\mainabltabheight}{0.088\linewidth}
    \hspace{-1em}
    \begin{subtable}{.3\linewidth}
        \centering
        \caption{Refiner Depth}
        \label{tab:main_ablations_refiner_depth}
        \adjustbox{max height=\mainabltabheight,max width=\linewidth}{\scalebox{.7}{
        \begin{tabular}{ccc}
            \toprule
            Refiner Depth ($d_r$) & {Params} & FID$\downarrow$ \\
            \midrule
            {\color{ourgray}--} & \phantom{+}{\color{ourgray}310M} & {\color{ourgray}3.30} \\
            0 & \phantom{00}+8M & 3.02 \\
            1 & \phantom{0}+27M & 2.85 \\
            \rowcolor{ourwhite} 2 & \phantom{0}+46M & 2.81 \\
            4 & \phantom{0}+84M & 2.82 \\
            8 & +160M & 2.81 \\
            \bottomrule
        \end{tabular}
        }}
    \end{subtable}\hspace{-3em}\hfill\hspace{-3em}
    \begin{subtable}{.51\linewidth}
        \centering
        \caption{Trainable Components}
        \label{tab:main_ablations_trainable_components}
        \adjustbox{max height=\mainabltabheight,max width=\linewidth}{\scalebox{.7}{
        \begin{tabular}{ccccc}
            \toprule
            \multicolumn{3}{c}{Trained?} & \multirow{2}{*}[-3pt]{\shortstack{Trainable\\Params}} & \multirow{2}{*}[-3pt]{FID$\downarrow$} \\
            \cmidrule{1-3}
            Output Head\  & \ Embedding\  & \ Base Backbone \\
            \midrule
            \multicolumn{1}{l}{\color{ourgray}{Baseline}} & & & {\color{ourgray}310M} & {\color{ourgray}3.30} \\
            \cmark & \xmark & \xmark & \phantom{0}46M & 2.86 \\
            \xmark & \cmark & \xmark & \phantom{0}40M & 2.85 \\
            \rowcolor{ourwhite} \cmark & \cmark & \xmark & \phantom{0}46M & 2.81 \\
            {\color{ourgray}\cmark} & {\color{ourgray}\cmark} & {\color{ourgray}\cmark} & {\color{ourgray}356M} & {\color{ourgray}2.72} \\
            \bottomrule
        \end{tabular}
        }}
    \end{subtable}\hspace{-3em}\hfill\hspace{-3em}
    \begin{subtable}{.30\linewidth}
        \centering
        \caption{Scales with Refiner}
        \label{tab:main_ablations_scales}
        \adjustbox{max height=\mainabltabheight,max width=\linewidth}{\scalebox{.7}{
        \begin{tabular}{lc}
            \toprule
            Scales & FID$\downarrow$ \\
            \midrule
            \rowcolor{ourwhite} all & 2.81 \\
            all $\setminus\; \{2^2\}$ & 2.98 \\
            all $\setminus\; \{3^2\}$ & 2.90 \\
            all $\setminus\; \{4^2\}$ & 2.86 \\
            all $\setminus\; \{5^2\}$ & 2.86 \\
            \multicolumn{2}{c}{\color{ourgray}[ctd. $\rightarrow$]} \\
            \bottomrule
        \end{tabular}
        \begin{tabular}{lc}
            \toprule
            Scales & FID$\downarrow$ \\
            \midrule
            all $\setminus\; \{6^2\}$ & 2.85 \\
            all $\setminus\; \{8^2\}$ & 2.81 \\
            all $\setminus\; \{10^2\}$ & 2.85 \\
            all $\setminus\; \{13^2\}$ & 2.86 \\
            all $\setminus\; \{16^2\}$ & 2.83 \\
            \color{ourgray} -- & \color{ourgray} 3.30 \\
            \bottomrule
        \end{tabular}
        }}
    \end{subtable}
    \hspace{-1em}
\end{table}

\paragraph{Small Refiners Suffice}
\Cref{tab:main_ablations_refiner_depth} varies the number of refiner transformer blocks ($d_r$).
Even a depth-0 refiner (no additional transformer blocks, just the autoregressive input projection and finetuned head with AR sampling) provides a meaningful improvement.
This demonstrates that the autoregressive factorization \emph{itself} is valuable, even when combined with just a causal context of one token and a single additional linear layer to incorporate extra information.
Performance saturates quickly, with $d_r = 2$ providing a favorable tradeoff.

\paragraph{Trainable Components}
\Cref{tab:main_ablations_trainable_components} ablates which components need to be trained.
Reusing the frozen input embedding or output head from the base VAR model provides a small performance regression compared to training the whole refiner jointly, indicating that the backbone's learned representations already carry the relevant information -- the refiner merely needs to model the residual dependencies that independent sampling discards.
Jointly training the VAR backbone yields only minor further gains (FID 2.72) at greatly increased training cost, so we keep the backbone frozen throughout.

\paragraph*{Which Scales Benefit Most?}
\Cref{tab:main_ablations_scales} measures each scale's contribution by applying the refiner at all but one scale during sampling.
The FID degradation relative to full-refiner sampling quantifies how much joint modeling at that scale matters for generation quality.
The largest drops occur at the earliest scales. This is notable, as the refiner's computational cost is also lowest at these scales, suggesting that selectively applying the refiner only at early scales could reduce inference overhead with minimal quality loss (\cref{sec:efficiency}).

Collectively, these ablations confirm that the \methodname's gains stem from restoring intra-scale dependencies, not from additional capacity or training, and that the module is robust across architectural choices, with efficient add-on training on a \emph{frozen pretrained} backbone sufficient to capture significant gains.

\begin{figure}[t]
    \centering
    \includegraphics[width=\linewidth]{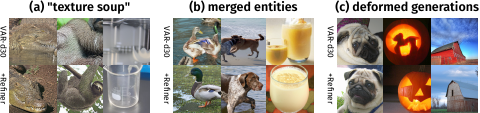}
    \caption{\textbf{VAR Failure Cases vs.\ Refiner.} Our \methodname can address a range of typical failures of VAR. Each column shows a paired sample (same class, same seed). Top: samples covering various failure modes from VAR-d30~\cite{tian2024visual}. Bottom: with refiner.}
    \label{fig:var_failures}
\end{figure}

\subsection{Main Results}
\paragraph{ImageNet Generation}
We identify three recurring failure modes of VAR that persist even at multi-billion parameter scales (\cref{fig:var_failures}-top):
samples that consist of class-relevant textures with little obvious structure (``texture soup''),
images with multiple inconsistent, often merged instances of the target class,
and samples with locally inconsistent structure.
All three stem from the lack of spatial coordination inherent in independent within-scale sampling (\cref{sec:method_var_prelim}).
Adding our refiner enables the model to sample from the joint intra-scale token distribution, directly addressing these failure modes (\cref{fig:var_failures}-bottom).
We show additional qualitative samples in Supp.~\cref{sec:app_qualitative_imagenet}.

\begin{table}[t]
    \centering
    \caption{\textbf{System-level Comparison} on class-conditional ImageNet-$256^2$ across discrete-token scale-wise autoregressive models. Within each backbone scale, VAR + Refiner achieves the best FID, improving by 0.16 to 0.49 over VAR with only $\sim$10\% additional parameters. See \cref{sec:app_additional_eval_details} for additional models \& evals w/o CFG.}
    \label{tab:imagenet_256}
    \newcommand{\improvement}[1]{$_{\color{ourgreen}\blacktriangledown #1}$}
    \newcommand{\improvementphantom}{\phantom{\improvement{0.00}}}
    \begin{subtable}{.49\linewidth}
        \centering
        \adjustbox{max width=\linewidth}{\scalebox{.8}{
        \begin{tabular}{lccccc}
            \toprule
            Method & Params & FID$\downarrow$ & IS$\uparrow$ & Prec$\uparrow$ & Rec$\uparrow$ \\
            \midrule
            MVAR-d16~\cite{zhang2026mvar} & 310M & \phantom{0}3.09\improvementphantom & 285.5 & 0.85 & 0.51 \\
            M-VAR-d16~\cite{ren2024m} & 464M & \phantom{0}\underline{3.07}\improvementphantom & 294.6 & 0.84 & 0.53 \\
            HMAR-d16~\cite{kumbong2025hmar} & 465M & \phantom{0}3.01\improvementphantom & 288.6 & 0.84 & 0.55 \\
            VAR-d16~\cite{tian2024visual} & 310M & \phantom{0}3.30\improvementphantom & 274.4 & 0.84 & 0.51 \\
            \rowcolor{ourwhite} + Refiner (Ours) & 356M & \phantom{0}\samebf{2.81}\improvement{0.49} & 267.2 & 0.81 & 0.56 \\
            \midrule
            HART-d20~\cite{tang2024hart} & 649M & \phantom{0}\underline{2.39}\improvementphantom & 316.4 & -- & -- \\
            MVAR-d20~\cite{zhang2026mvar} & 600M & \phantom{0}2.87\improvementphantom & 295.3 & 0.86 & 0.52 \\
            M-VAR-d20~\cite{ren2024m} & 900M & \phantom{0}2.41\improvementphantom & 308.4 & 0.85 & 0.58 \\
            HMAR-d20~\cite{kumbong2025hmar} & 840M & \phantom{0}2.50\improvementphantom & 319.0 & 0.85 & 0.57 \\
            VAR-d20~\cite{tian2024visual} & 600M & \phantom{0}2.57\improvementphantom & 302.6 & 0.83 & 0.56 \\
            \rowcolor{ourwhite} + Refiner (Ours) & 671M & \phantom{0}\samebf{2.17}\improvement{0.40} & 274.7 & 0.80 & 0.60 \\
            \midrule
            \color{ourgray} ImageNet Validation & \color{ourgray} -- & \color{ourgray} \phantom{0}1.78\improvementphantom & \color{ourgray} -- & \color{ourgray} -- & \color{ourgray} -- \\
            \bottomrule
        \end{tabular}
        }}
    \end{subtable}\hfill
    \begin{subtable}{.49\linewidth}
        \centering
        \adjustbox{max width=\linewidth}{\scalebox{.8}{
        \begin{tabular}{lccccc}
            \toprule
            Method & Params & FID$\downarrow$ & IS$\uparrow$ & Prec$\uparrow$ & Rec$\uparrow$ \\
            \midrule
            HART-d24~\cite{tang2024hart} & 1.0B & \phantom{0}2.00\improvementphantom & 331.5 & -- & -- \\
            FastVAR-d24~\cite{guo2025fastvar} & 1.0B & \phantom{0}2.64\improvementphantom & 287.4 & 0.80 & 0.58 \\
            MVAR-d24~\cite{zhang2026mvar} & 1.0B & \phantom{0}2.23\improvementphantom & 300.1 & 0.86 & 0.52 \\
            M-VAR-d24~\cite{ren2024m} & 1.5B & \phantom{0}\underline{1.93}\improvementphantom & 320.7 & 0.83 & 0.59 \\
            HMAR~\cite{kumbong2025hmar} & 1.3B & \phantom{0}2.10\improvementphantom & 319.0 & 0.83 & 0.60 \\
            VAR-d24~\cite{tian2024visual} & 1.0B & \phantom{0}2.09\improvementphantom & 312.9 & 0.83 & 0.57 \\
            \rowcolor{ourwhite} + Refiner (Ours) & 1.1B & \phantom{0}\samebf{1.83}\improvement{0.26} & 288.2 & 0.79 & 0.63 \\
            \midrule
            HART-d30~\cite{tang2024hart} & 2.0B & \phantom{0}\underline{1.77}\improvementphantom & 330.3 & -- & -- \\
            FastVAR-d30~\cite{guo2025fastvar} & 2.0B & \phantom{0}2.30\improvementphantom & 288.7 & 0.81 & 0.59 \\
            VAR-CoDe-d30~\cite{chen2025collaborative} & 2.3B & \phantom{0}1.94\improvementphantom & 296\phantom{.0} & 0.81 & 0.60 \\
            HMAR~\cite{kumbong2025hmar} & 2.4B & \phantom{0}1.95\improvementphantom & 334.5 & 0.82 & 0.62 \\
            VAR-d30~\cite{tian2024visual} & 2.0B & \phantom{0}1.92\improvementphantom & 323.1 & 0.82 & 0.58 \\
            \rowcolor{ourwhite} + Refiner (Ours) & 2.2B & \phantom{0}\samebf{1.76}\improvement{0.16} & 319.4 & 0.80 & 0.62 \\
            \midrule
            M-VAR-d32~\cite{ren2024m} & 3.0B & \phantom{0}1.78\improvementphantom & 331.2 & 0.83 & 0.61 \\
            \bottomrule
        \end{tabular}
        }}
    \end{subtable}
\end{table}

\begin{figure}[t]
    \centering
    \includegraphics[width=.9\linewidth]{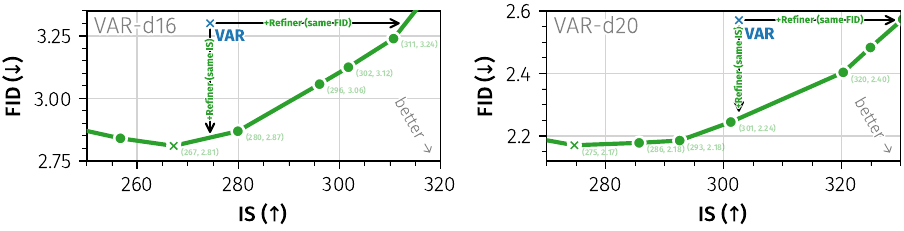}
    \caption{\textbf{FID-IS Improvement Tradeoff.} By varying the classifier-free guidance scale, the logit refiner can achieve improvements in both dimensions.}
    \label{fig:fid_is}
\end{figure}

These qualitative gains are reflected in quantitative metrics.
\Cref{tab:imagenet_256} compares our \methodname applied to VAR across scales and with a broad range of scale-wise autoregressive methods.
Reference results for other generative model families are reported in the extended comparison (\cref{tab:imagenet_256_extended}).
Across all model scales, the refiner consistently improves FID by a significant margin (0.16 to 0.49), while only adding $\sim$10\% additional parameters.
VAR-d24 + Refiner (1.1B params total) even exceeds VAR-d30 (2B params) by a significant margin, obtaining a stronger model at roughly half the size.
Beyond FID, the refiner also consistently improves recall by 0.04 to 0.06 across all backbone scales, indicating that restoring intra-scale dependencies recovers sample diversity rather than trading it away.
The small accompanying decrease in Inception Score follows from the lower CFG scales that are FID-optimal for the refiner, not from reduced sample quality -- by varying the guidance scale, improvements in FID and/or IS over the baseline can be traded off (see \cref{fig:fid_is}).
We also compare in a CFG-free setting in \cref{tab:unguided}, where the logit refiner also consistently outperforms the baseline.
Compared with other VAR variants that address orthogonal aspects, the basic VAR model with our refiner achieves the best generative performance within each model scale.
Our approach does not utilize any of the improvements introduced by these methods, which may lead to further gains in combination. These directions are left for future work.

\begin{figure}[t]
    \centering
    \includegraphics[width=\linewidth]{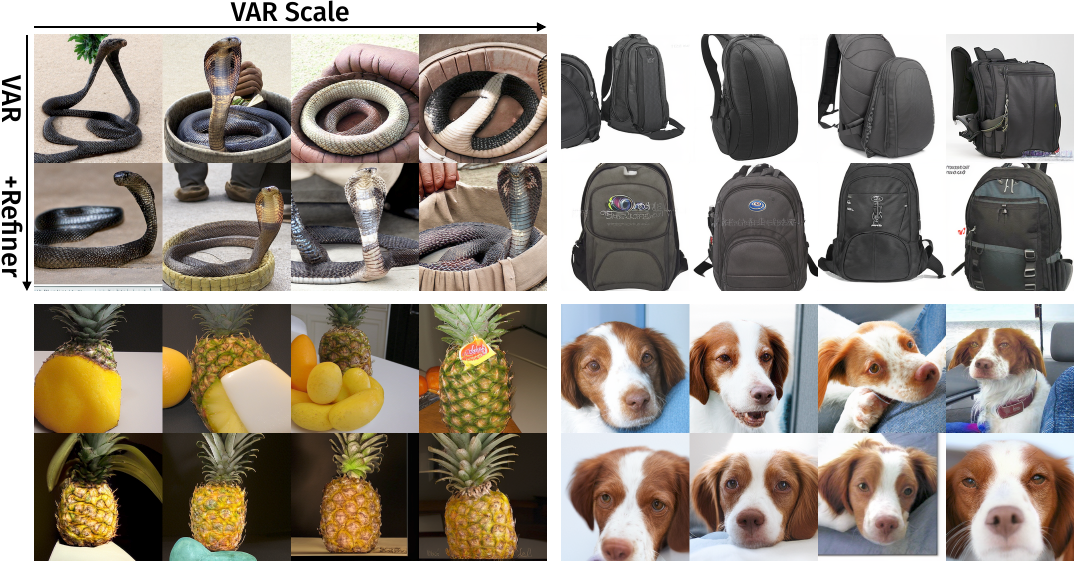}
    \caption{\textbf{Qualitative Scaling Behavior.} Paired samples (same class, seed) across backbone scales (VAR-\{16, 20, 24, 30\}). \emph{Top rows:} vanilla VAR improves visual fidelity and class consistency with scale, but spatial consistency problems persist at every size. \emph{Bottom rows:} adding our refiner resolves these artifacts across all scales.}
    % \vspace{-3em}
    \label{fig:refiner_scaling}
\end{figure}

\paragraph{Scaling Behavior}
\Cref{fig:refiner_scaling} investigates how the \methodname interacts with backbone scale.
Qualitatively, (\cref{fig:refiner_scaling}), spatial consistency problems persist across all vanilla VAR scales, even as visual fidelity improves with model size.
The refiner resolves these artifacts at every scale, confirming that the underlying issue is the mean-field-style sampling rule rather than insufficient model capacity.
Quantitatively (\cref{fig:teaser}b), the refiner shifts the FID scaling curve downward across all backbone sizes without altering the overall scaling trend, reflecting the qualitative scaling findings, and demonstrating that correcting the mean-field-style approximation can be more parameter-efficient than scaling the backbone.

\paragraph{Training and Inference Efficiency}\label{sec:efficiency}
The \methodname corrects a sampling-time approximation, modeling \emph{residual} dependencies between tokens within each scale.
This suggests that a refiner trained on a frozen backbone should already capture most of the achievable gains, since the marginals are correct and only the dependencies are missing.
Our results confirm this:
adding a refiner to a frozen backbone improves the FID from 3.30 to 2.81 with only 66 H200-h of additional training compute.
Training the full model with an integrated refiner from scratch yields a stronger FID of 2.57, but requires 1{,}845 H200-h -- 28$\times$ more compute for the remaining third of improvement.
Jointly finetuning the backbone occupies a middle ground (FID 2.72, 127 H200-h).
The dominant effect is thus the correction of the mean-field-style factorization itself, achievable as a lightweight post-hoc addition to any pretrained VAR model without retraining the backbone.

During inference, the refiner introduces sequential within-scale sampling, but the cost is modest.
The expensive backbone forward pass remains fully parallel:
for each scale $k$, the backbone computes all hidden states $\mathbf{h}^{(k)}$ in a single pass.
Only the lightweight refiner blocks ($d_r = 2$ blocks vs.\ $d \in [16, 30]$ backbone blocks) run sequentially, and we employ KV caching to avoid redundant computations across tokens within a scale.

\begin{figure}[t]
    \centering
    \begin{subfigure}[b]{.31\linewidth}
        \centering
        \includegraphics[width=\linewidth]{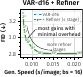}
        \caption{VAR-d16 stage tradeoff}
        \label{fig:var_tradeoff}
    \end{subfigure}\hfill
    \begin{subfigure}[b]{.66\linewidth}
        \centering
        \includegraphics[width=.9\linewidth]{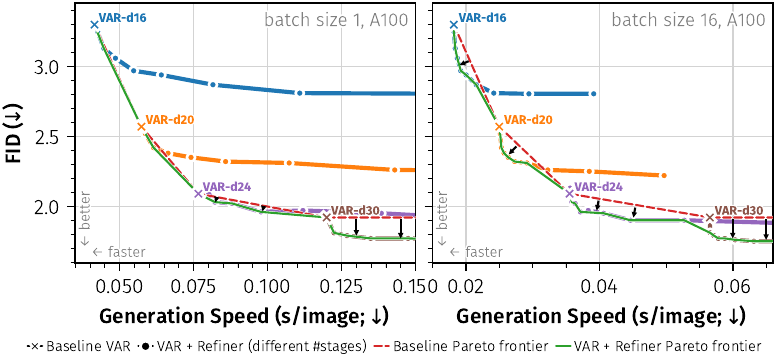}
        \caption{Across backbone scales and batch sizes}
        \label{fig:performance}
    \end{subfigure}
    \caption{\textbf{Quality/Efficiency Tradeoff.} \textbf{(a)} For VAR-d16, applying the refiner only at the first $k$ scales (\cref{tab:main_ablations_scales}) traverses the quality/efficiency tradeoff. \textbf{(b)} The same effect across backbone scales (d\{16,20,24,30\}) and batch-size regimes. \textit{Left:} latency-optimized regime (batch size 1, the worst case for the refiner); \textit{Right:} typical regime (batch size 16). At batch size 16, the refiner Pareto-dominates the baseline at every scale; at batch size 1, it adds intermediate operating points at low depths and dominates from d24/d30.}
    \label{fig:efficiency}
\end{figure}

The relative overhead remains modest across backbone sizes and batch sizes (\cref{fig:performance}), since the expensive backbone still runs once per scale in parallel and only the two-layer refiner is sequential -- the refiner does not turn VAR into a fully token-wise autoregressive model.
At a typical batch size, the refiner improves the quality/efficiency Pareto frontier over vanilla VAR at \emph{every} backbone scale; in the latency-optimized single-sample regime, it adds finer-grained operating points and dominates the baseline from VAR-d24 onward.

Moreover, the scales ablation (\cref{tab:main_ablations_scales}) shows that the refiner's quality gains concentrate at early scales, which contain the fewest tokens.
For VAR-d16, this enables a practical trade-off (\cref{fig:var_tradeoff}):
applying the refiner only at the first few scales can substantially reduce the sequential sampling cost while retaining most of the quality improvement.
Concretely, applying the refiner on scales up to $8^2$/$10^2$ reduces the refiner overhead by 84\%/71\% while retaining 88\%/99\% of the full FID improvement, respectively.

\begin{figure}[t]
    \centering
    \includegraphics[width=\linewidth]{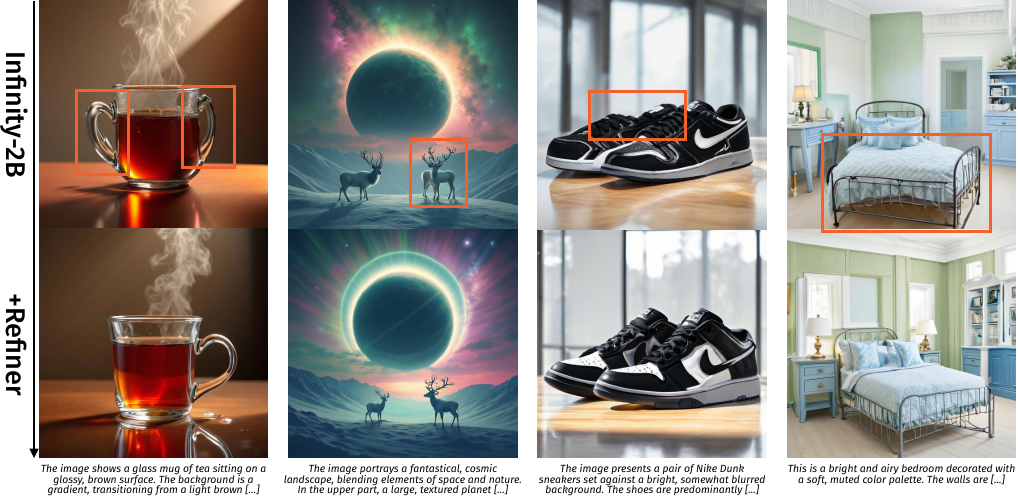}
    \caption{\textbf{Qualitative Text-to-Image Results.} We show results from Infinity-2B~\cite{han2025infinity} at $1024^2$ resolution without (top) and with our refiner (bottom). As in the ImageNet class-conditional case, our \methodname improves spatial coherence (inconsistencies in original images marked in {\color{ourorange}\textbf{orange}}) of the generated images.}
    \label{fig:t2i_qualitative}
    % \vspace{-1em}
\end{figure}

\begin{table}[t]
    \centering
    \caption{\textbf{Quantitative T2I Results on HPSv3~\cite{ma2025hpsv3}.} We evaluate automated preference scores (higher is better) for samples generated with and without our \methodname. It improves spatial consistency in generated images, reflected in better scores.}
    \newcommand{\rotheada}[1]{\rotatebox{65}{\makebox[0pt][l]{\scriptsize #1}}}
    \adjustbox{max width=\linewidth}{
    \begin{tabular}{l>{\color{ourgray}}c>{\color{ourgray}}c>{\color{ourgray}}c>{\color{ourgray}}c>{\color{ourgray}}c>{\color{ourgray}}c>{\color{ourgray}}c>{\color{ourgray}}c>{\color{ourgray}}c>{\color{ourgray}}c>{\color{ourgray}}c>{\color{ourgray}}cc}
    \toprule
    Model 
    & \rotheada{Animals} \rule{0pt}{14ex}
    & \rotheada{Architecture} 
    & \rotheada{Arts} 
    & \rotheada{Characters} 
    & \rotheada{Design} 
    & \rotheada{Food} 
    & \rotheada{Natural Scenery} 
    & \rotheada{Others} 
    & \rotheada{Plants} 
    & \rotheada{Products} 
    & \rotheada{Science} 
    & \rotheada{Transportation} 
    & Average$\uparrow$ \\
    \midrule
    Infinity-2B~\cite{han2025infinity} & \samebf{9.57} & 10.03 & 9.81 & 11.10 & 9.41 & 10.57 & \samebf{9.08} & 10.13 & \samebf{10.14} & 9.61 & \samebf{8.57} & 9.48 & 9.79 \\
    + Refiner ($d_r$ = 2) & 9.53 & \samebf{10.20} & \samebf{10.02} & \samebf{11.29} & \samebf{9.68} & \samebf{10.73} & 9.04 & \samebf{10.18} & \samebf{10.14} & \samebf{9.85} & 8.50 & \samebf{9.72} & \samebf{9.91} \\
    \bottomrule
\end{tabular}
    }
    % \vspace{-1em}
    \label{tab:hps}
\end{table}

\paragraph{Scaling to T2I}
To validate that the \methodname generalizes beyond class-conditional ImageNet, we apply it to Infinity~\cite{han2025infinity}, a scaled VAR variant for text-to-image synthesis.
We train the refiner on Infinity's backbone following a similar setup as for VAR (frozen backbone, $d_r = 2$, 100k steps at batch size 768; $\sim$640 H200-h train time -- orders of magnitude less than the base model's pretraining time) on images and captions from FLUX-6M~\cite{fang2025flux}. % 2 * 3.84e7/2130/3600*(768/12)
Following the findings from the previous paragraph, we apply the refiner selectively to the first several stages (up to resolution $6^2$).
Qualitatively (\cref{fig:t2i_qualitative}; see also Supp.~\cref{sec:app_qualitative_t2i} for additional examples), the refiner yields the same types of improvements observed on ImageNet:
improved structural coherence and reduced texture inconsistencies.
These improvements are also reflected in quantitative evaluations: the refiner improves HPSv3~\cite{ma2025hpsv3} scores from 9.79 to 9.91 (see \cref{tab:hps}), confirming that the benefits of restoring intra-scale dependencies transfer to open-vocabulary text-to-image generation.

\section{Conclusion}
% We have identified a fundamental limitation of scale-wise visual autoregressive generation: parallel within-scale decoding constitutes a mean-field-style approximation that discards spatial dependencies among tokens, causing locally incoherent samples regardless of backbone capacity or training duration.
Scale-wise visual autoregressive generation has a fundamental limitation: parallel within-scale decoding constitutes a mean-field-style approximation that discards spatial dependencies among tokens, causing locally incoherent samples regardless of backbone capacity or training duration.
The \methodname addresses this by restoring intra-scale dependencies through a lightweight autoregressive module that operates on parallel backbone features, adding only $\sim$10\% parameters and requiring less than 5\% of the base model's training compute.
Across a large range of backbone sizes and both class- and text-conditional generation, the refiner consistently improves generation quality by a wide margin.
Controlled ablations isolate joint intra-scale sampling rather than additional capacity or training as the critical ingredient.

\paragraph{Limitations and Future Work}
    Autoregressive within-scale sampling introduces sequential computation at each scale.
    While applying the refiner selectively at early scales retains the vast majority of quality gains at a fraction of the cost, the overhead is not fully eliminated.
More broadly, our results suggest that when parallel decoding introduces mean-field-style assumptions, lightweight autoregressive correction can restore the discarded dependencies at minimal cost
% -- a principle that may extend beyond VAR to other parallel generative architectures.
-- a potentially general principle that shows promise for applications in parallel generative architectures~\citep[e.g.][]{chang2022maskgit,li2024autoregressive}.
    Combining the refiner with orthogonal VAR improvements~\citep[e.g.][]{tang2024hart,ren2024m,zhang2026mvar} and exploring non-autoregressive approaches to within-scale sampling are further promising directions.

% \omitforsubmission{
%     \section*{Acknowledgment}
%     Jack Ghallager\stefan{check with him}, Ulrich Prestel, Tommaso Martorella, Ming Gui, Nick Stracke, Kolja Bauer, and Vincent Tao Hu for helpful discussions and advice.
% }

\section*{Acknowledgments}
This project has been supported by the Horizon Europe project ELLIOT (GA No.\ 101214398), the project ``GeniusRobot'' (01IS24083) funded by the Federal Ministry of Research, Technology and Space (BMFTR), the BMWE ZIM-project (No.\ KK5785001LO4) ``conIDitional LoRA'', the German Federal Ministry for Economic Affairs and Energy within the project ``NXT GEN AI METHODS - Generative Methoden für Perzeption, Prädiktion und Planung'', and the bidt project KLIMA-MEMES. The authors gratefully acknowledge the Gauss Center for Supercomputing for providing compute through the NIC on JUWELS/JUPITER at JSC and the HPC resources supplied by the NHR@FAU Erlangen.
We thank Ulrich Prestel, Jack Gallagher, Tommaso Martorella, Ming Gui, Nick Stracke, Kolja Bauer, and Vincent Tao Hu for feedback, proofreading, and helpful discussions, and Owen Vincent for technical support.

% ---- Bibliography ----
%
% BibTeX users should specify bibliography style 'splncs04'.
% References will then be sorted and formatted in the correct style.
%
\bibliographystyle{splncs04}
\bibliography{main}

\clearpage
\setcounter{page}{1}
\section*{Supplementary Material}

\setcounter{figure}{0}
\setcounter{table}{0}
\setcounter{equation}{0}
\setcounter{section}{0}
\renewcommand\thesection{\Alph{section}}
\renewcommand\thefigure{\Alph{section}.\arabic{figure}}
\renewcommand\thetable{\Alph{section}.\arabic{table}}
\renewcommand\theequation{\Alph{section}.\arabic{equation}}

\crefalias{section}{appsec}
\crefalias{figure}{appfig}
\crefalias{table}{apptab}
\crefalias{equation}{appeq}

% \omitforsubmission{
%     \section*{Notes}
%     \paragraph{Implementation Notes}
%     \begin{itemize}
%         \item My (Stefan) current implementation uses ImageNet shards. We should, for optimal performance, consider mapping the full ImageNet to a ramdisk (once per node via node-wide mutex) at startup once (ImageNet is ~150GB, so too much to do once per rank, but once per node will be fine, even on Jupiter), just to ensure that we \textit{definitely} don't have any sampling biases
%     \end{itemize}
% }

\section{Implementation Details}\label{sec:app_implementation_details}

\paragraph{Training Hyperparameters}
We show relevant hyperparameters for all trained model variations in \Cref{tab:hparams}.
Optimizer settings have been directly taken from VAR, with the exception of a reduced learning rate (result of a sweep; we find that the learning rate does not meaningfully influence the FID once converged for our base configuration, with more than double and less than half the learning rate resulting in similar sample metrics) and a linear warmup + cosine decay schedule.
% For ablations, we train up to 200k steps (40 epochs), taking approx.\ 7h. All variants are consistently converged by that time.
For all VAR models, we train for 40 epochs. All variants are converged by that time. For Infinity~\cite{han2025infinity}, we train for 200k steps.

\begin{table}[H]
    \centering
    \caption{\textbf{Hyperparameters.}}
    \adjustbox{max width=\linewidth}{
    \begin{tabular}{lccc}
        \toprule
        Variant & Ablations & Main Results & Text-to-Image \\
        \midrule
        Dataset & ImageNet-$256^2$ & ImageNet-$256^2$ & FLUX-6M~\cite{fang2025flux} \\
        \midrule
        Base Model & VAR-d16~\cite{tian2024visual} & VAR-d\{16,20,24,30\}~\cite{tian2024visual} & Infinity-2B~\cite{han2025infinity} \\
        Base Model Depth $d$ & 16 & \{16,20,24,30\} & 32\\
        Base Model Width & 1024 & \{1024,1280,1536,1920\} & 2048 \\
        \midrule
        Refiner Depth $d_r$ & \{0, 1, \underline{2}, 4, 8\} & 2 & 2 \\
        Refiner Width & 1024 (matching base) & matching base & 2048 (matching base) \\
        Trainable Parameters & \{8M,27M,\underline{46M},84M,160M\} & varying & 193M \\
        \midrule
        Batch Size & 768 & 768 if $d < 30$, else 1024 & 768 \\
        Training Duration & 30 epochs & 30 epochs & 200k steps \\
        \midrule
        Precision & fp16 MP & fp16 MP & bf16 MP \\
        Training Hardware & 8 H200 & 8-32 H200 & 64 H200 \\
        % Step Time & 0.13+s \\
        \midrule
        Optimizer & AdamW~\cite{loshchilov2017decoupled} & AdamW~\cite{loshchilov2017decoupled} & AdamW~\cite{loshchilov2017decoupled} \\
        Base LR ($LR = LR_\text{base} \cdot BS/256$) & $2.5\cdot 10^{-5}$ & $2.5\cdot 10^{-5}$ & $2.5\cdot 10^{-5}$ \\
        Learning Rate Warmup & 2\% of training from 0.5\% of peak & 2\% of training from 0.5\% of peak & 2\% of training from 0.5\% of peak \\
        % Learning Rate Warmup & \multicolumn{3}{c}{2\% of training from 0.5\% of peak} \\
        Learning Rate Schedule & linear & linear & linear \\
        Betas $(\beta_1, \beta_2)$ & $(0.9, 0.95)$ & $(0.9, 0.95)$ & $(0.9, 0.95)$ \\
        Weight Decay & 0.05 & 0.05; d30: scheduled following VAR~\cite{tian2024visual} & 0.01 \\
        \bottomrule
    \end{tabular}
    }
    \label{tab:hparams}
\end{table}

\begin{figure}[t]
    \centering
    \includegraphics[width=\linewidth]{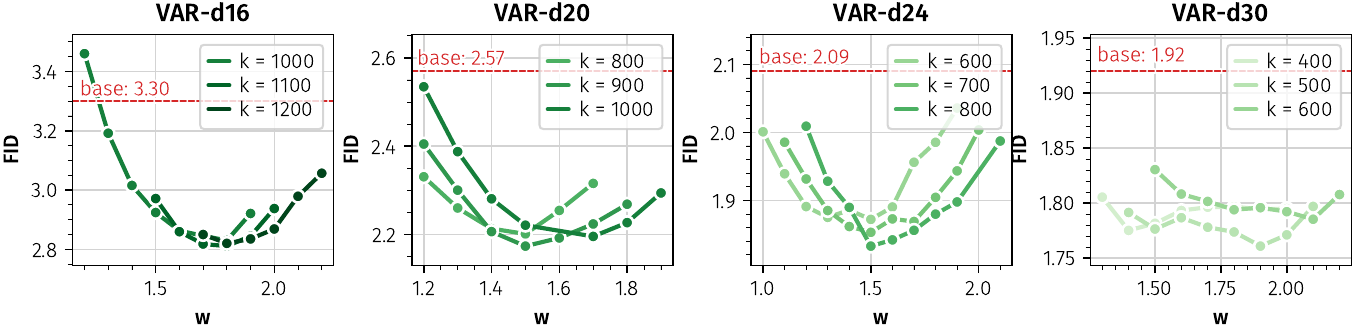}
    \caption{\textbf{Inference Hyperparameter Exploration.} The optimal classifier-free guidance scale $w$ and top-$k$ threshold for VAR with our refiner vary across scales.}
    \label{fig:fid_sweep}
\end{figure}

\paragraph{Inference Hyperparameters}
We observe that, like with most other image generation models, final performance of VAR + \methodname as measured by FID varies with inference hyperparameters, whose optimal values vary with base model size.
We therefore sweep both top-k and CFG~\cite{ho2022classifier} scales individually w.r.t.\ FID.
Over CFG, results are generally smooth (i.e., straightforward to sweep, typically close to convex).
\Cref{fig:fid_sweep} shows the result of our hyperparameter search, where we sweep the CFG scale in increments of 0.1 and $k$ in increments of 100 around the optimum.
FID~\cite{heusel2017gans} is computed on 50k samples with class-balanced sampling following the baseline VAR~\cite{tian2024visual}.
The optimal params we found are listed in \cref{tab:inference_hparams}.

\begin{table}[t]
    \centering
    \caption{\textbf{Inference Hyperparameters.} Used for quantitative evaluations. Parameters for Infinity-2B directly mirror those of the base model.}
    \label{tab:inference_hparams}
    \begin{tabular}{lccccc}
        \toprule
        Base Model & VAR-d16 & VAR-d20 & VAR-d24 & VAR-d30 & Infinity-2B \\
        \midrule
        CFG $w$ & 1.8 & 1.5 & 1.5 & 1.9 & 3.0 \\
        Top-$k$ & 1100 & 900 & 800 & 500 & 900 \\
        Top-$p$ & -- & -- & -- & -- & 0.97 \\
        $\tau$ & -- & -- & -- & -- & 0.5 \\
        \bottomrule
    \end{tabular}
\end{table}

\paragraph{Evaluation Details}
    For our VAR-d16+2 refiner, we try training with different random seeds to estimate confidence intervals for our results.
    Across four training runs, we obtain the following FIDs: 2.77, 2.80, 2.81, 2.83, resulting in a sample standard deviation of 0.025.
    This puts our gains compared to all baselines in \cref{tab:imagenet_256} to at least 8 standard deviations, indicating that the gains are statistically significant.
    Across all runs, we do not choose specific checkpoints, but use the same training setup, including seed.
For HPSv3, we evaluate the first 50 prompts per subset (600 images total), since sampling all images would be prohibitively expensive.

\begin{table}[t]
    \centering
    \caption{\textbf{Influence of Intra-Scale Token Ordering.} All orderings yield near-identical gains.}
    \newcommand{\tokenorderimg}[1]{\includegraphics[width=2.5em]{img/token_orders_#1.pdf}}
    \newcommand{\improvement}[1]{{$_{\color{ourgreen}\blacktriangledown #1}$}}
    \adjustbox{max width=.6\linewidth}{
    \begin{tabular}{l@{\hskip 1.2em}cccccc}
        \toprule
        Method & VAR-d16 & \multicolumn{5}{c}{+ Refiner (Ours)} \\
        \cmidrule(lr){2-2} \cmidrule(lr){3-7}
        \multirow{2}{*}[-1em]{\shortstack{Sampling\\Order}} & parallel & sweep (paper) & col.-major & alternate & spiral in & spiral out \\
        & \tokenorderimg{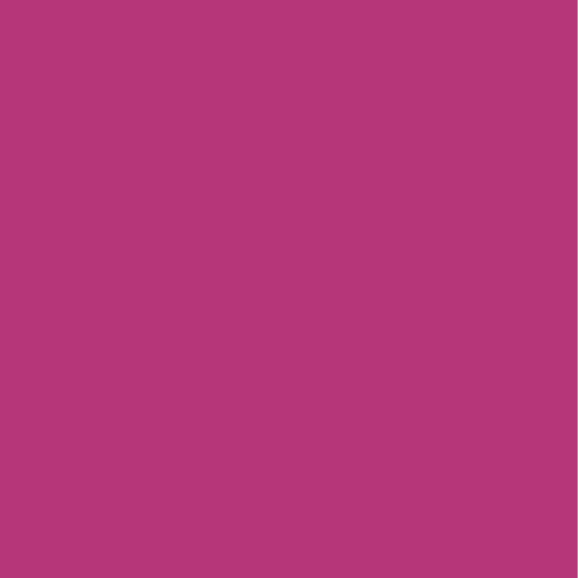} & \tokenorderimg{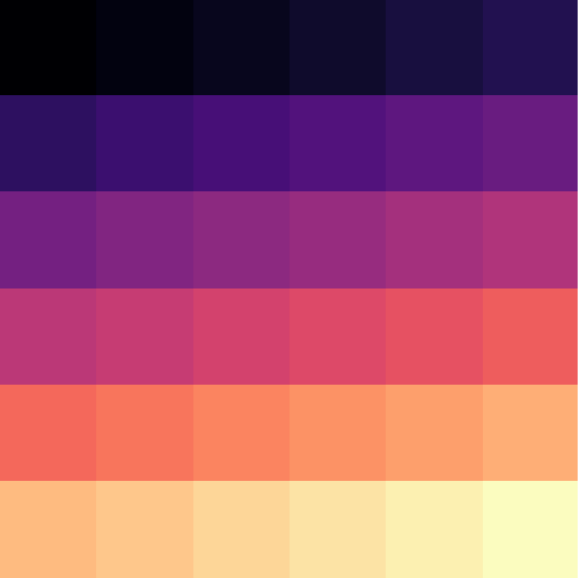} & \tokenorderimg{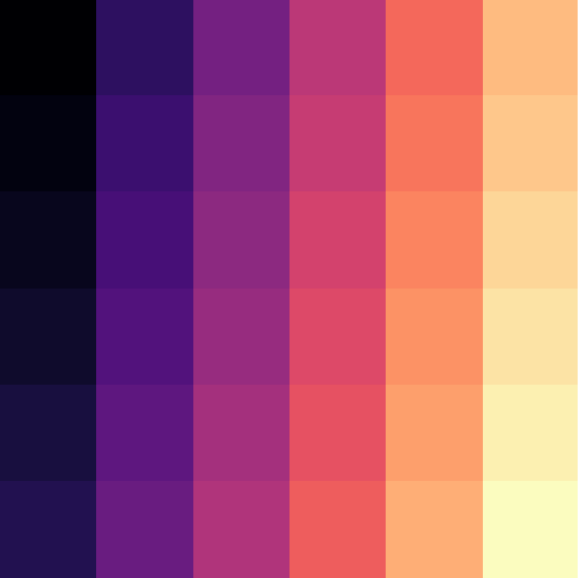} & \tokenorderimg{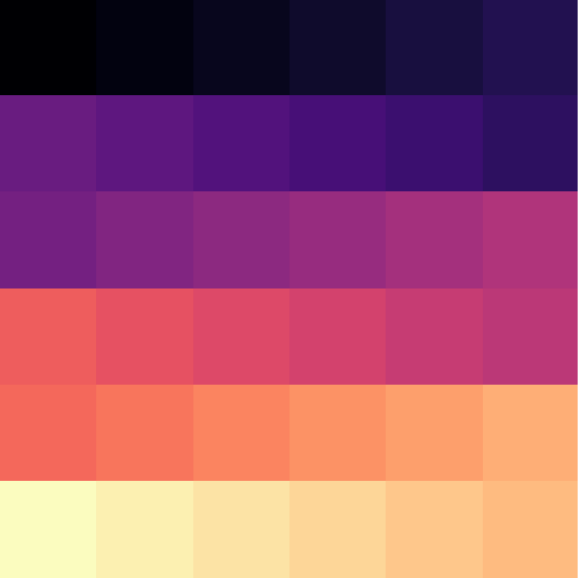} & \tokenorderimg{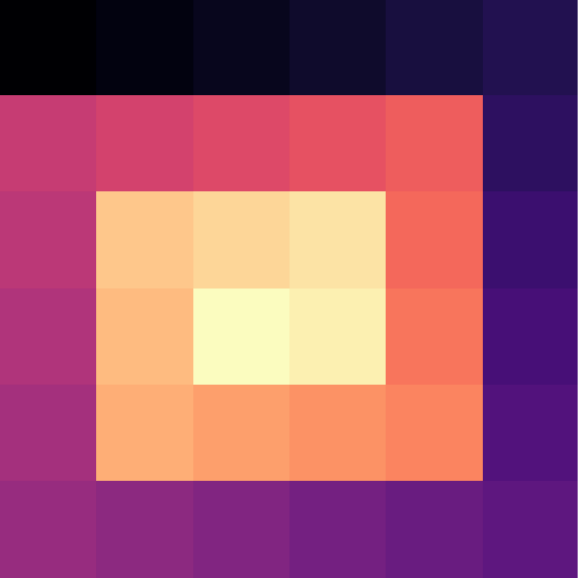} & \tokenorderimg{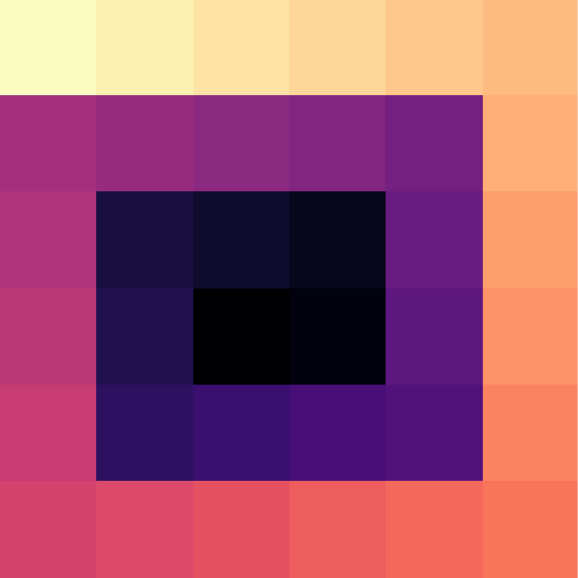} \\
        \midrule
        FID ($\downarrow$) & 3.30 & 2.81\improvement{0.49} & 2.83\improvement{0.47} & 2.82\improvement{0.48} & 2.83\improvement{0.47} & 2.86\improvement{0.44} \\
        \bottomrule
    \end{tabular}
    }
    \label{tab:token_order}
\end{table}

\paragraph{Within-Scale Token Ordering}
We generally use ``sweep''/row-major token ordering for the refiner.
We explore whether this choice is relevant for final performance by training refiners with five different intra-scale orderings in \cref{tab:token_order}.
Inference hparams are reused from original ``sweep'' order; per-order tuning would likely close the small remaining differences.
All tested orderings retain the gain, showing that the improvement is not an artifact of raster scan.

\section{Additional Quantitative Evaluation Details}
\label{sec:app_additional_eval_details}
We show an extended version of \cref{tab:imagenet_256} in \cref{tab:imagenet_256_extended}.
In addition to the major improvement in FID discussed in the main paper, we also observe a minor decrease in IS and precision, and an increase in recall.
% \TODO{Interpretation}
%     We interpret these results as a slight decrease in diversity of samples, which is directly explained by the addition of joint sampling:
%     generally, the support of the joint $p(a,b,\ldots)$ of random variables $a, b, \ldots$ is always larger than or equal to $p(a)\cdot p(b)\cdot \ldots$:
%     \begin{equation}
%         \mathrm{supp}(p(a)\cdot p(b)\cdot \ldots) \subseteq \mathrm{supp}(p(a, b, \ldots)).
%     \end{equation}
%     Thus, the support of the base VAR model's sample distribution will be a superset of the joint that our refiner tries to approximate, leading to higher diversity (but lower correctness, as this increase in support/diversity comes from samples that are \emph{not part of the joint}) for the base model, matching our quantitative observations.
    We attribute the reduction of IS values primarily to VAR with a refiner requiring lower CFG scales $w$ for optimal FID (higher guidance scales are generally associated with higher IS values).
    Further, we show evaluations without classifier-free guidance in \cref{tab:unguided}: gains with the refiner vs.~the baseline persist, and are more pronounced for large models than for small models.

\begin{table}[H]
    \centering
    \caption{\textbf{Extended System-level Comparison} on class-conditional ImageNet-$256^2$ (extending \cref{tab:imagenet_256}).}
    \label{tab:imagenet_256_extended}
    \newcommand{\improvement}[1]{$_{\color{ourgreen}\blacktriangledown #1}$}
    \newcommand{\improvementphantom}{\phantom{\improvement{0.00}}}
    \centering
    % \adjustbox{max width=\linewidth}{\scalebox{.85}{
    \adjustbox{max width=\linewidth}{\scalebox{.7}{
    \begin{tabular}{lccccc}
        \toprule
        Method & Params & FID$\downarrow$ & IS$\uparrow$ & Prec$\uparrow$ & Rec$\uparrow$ \\
        \midrule
        \multicolumn{4}{l}{\textbf{\textit{Scale-wise Autoregression~\cite{tian2024visual}}}} \\
        % \multicolumn{4}{l}{\TODO{other VAR follow-ups}} \\
        % \multicolumn{4}{l}{\TODO{FlowAR~\cite{ren2024flowar} seems like a relevant comparison}} \\
        MVAR-d16~\cite{zhang2026mvar} & 310M & \phantom{0}3.09\improvementphantom & 285.5 & 0.85 & 0.51 \\
        M-VAR-d16~\cite{ren2024m} & 464M & \phantom{0}\underline{3.07}\improvementphantom & 294.6 & 0.84 & 0.53 \\
        HMAR-d16~\cite{kumbong2025hmar} & 465M & \phantom{0}3.01\improvementphantom & 288.6 & 0.84 & 0.55 \\
        VAR-d16~\cite{tian2024visual} & 310M & \phantom{0}3.30\improvementphantom & 274.4 & 0.84 & 0.51 \\
        \rowcolor{ourwhite} + Refiner (Ours) & 356M & \phantom{0}\samebf{2.81}\improvement{0.49} & 267.2 & 0.81 & 0.56 \\
        \midrule
        HART-d20~\cite{tang2024hart} & 649M & \phantom{0}\underline{2.39}\improvementphantom & 316.4 & -- & -- \\
        MVAR-d20~\cite{zhang2026mvar} & 600M & \phantom{0}2.87\improvementphantom & 295.3 & 0.86 & 0.52 \\
        M-VAR-d20~\cite{ren2024m} & 900M & \phantom{0}2.41\improvementphantom & 308.4 & 0.85 & 0.58 \\
        HMAR-d20~\cite{kumbong2025hmar} & 840M & \phantom{0}2.50\improvementphantom & 319.0 & 0.85 & 0.57 \\
        VAR-d20~\cite{tian2024visual} & 600M & \phantom{0}2.57\improvementphantom & 302.6 & 0.83 & 0.56 \\
        \rowcolor{ourwhite} + Refiner (Ours) & 671M & \phantom{0}\samebf{2.17}\improvement{0.40} & 274.7 & 0.80 & 0.60 \\
        % \bottomrule
        % \toprule
        % Method & Params & FID$\downarrow$ & IS$\uparrow$ & Prec$\uparrow$ & Rec$\uparrow$ \\
        \midrule
        % \multicolumn{4}{l}{\textbf{\textit{Scale-wise Autoregression (ctd.)}}} \\
        HART-d24~\cite{tang2024hart} & 1.0B & \phantom{0}2.00\improvementphantom & 331.5 & -- & -- \\ 
        FastVAR-d24~\cite{guo2025fastvar} & 1.0B & \phantom{0}2.64\improvementphantom & 287.4 & 0.80 & 0.58 \\
        MVAR-d24~\cite{zhang2026mvar} & 1.0B & \phantom{0}2.23\improvementphantom & 300.1 & 0.86 & 0.52 \\
        M-VAR-d24~\cite{ren2024m} & 1.5B & \phantom{0}\underline{1.93}\improvementphantom & 320.7 & 0.83 & 0.59 \\
        HMAR~\cite{kumbong2025hmar} & 1.3B & \phantom{0}2.10\improvementphantom & 319.0 & 0.83 & 0.60 \\
        VAR-d24~\cite{tian2024visual} & 1.0B & \phantom{0}2.09\improvementphantom & 312.9 & 0.83 & 0.57 \\
        \rowcolor{ourwhite} + Refiner (Ours) & 1.1B & \phantom{0}\samebf{1.83}\improvement{0.26} & 288.2 & 0.79 & 0.63  \\
        \midrule
        HART-d30~\cite{tang2024hart} & 2.0B & \phantom{0}\underline{1.77}\improvementphantom & 330.3 & -- & -- \\
        FastVAR-d30~\cite{guo2025fastvar} & 2.0B & \phantom{0}2.30\improvementphantom & 288.7 & 0.81 & 0.59 \\
        VAR-CoDe-d30~\cite{chen2025collaborative} & 2.3B & \phantom{0}1.94\improvementphantom & 296\phantom{.0} & 0.81 & 0.60 \\
        HMAR~\cite{kumbong2025hmar} & 2.4B & \phantom{0}1.95\improvementphantom & 334.5 & 0.82 & 0.62 \\
        VAR-d30~\cite{tian2024visual} & 2.0B & \phantom{0}1.92\improvementphantom & 323.1 & 0.82 & 0.58 \\
        \rowcolor{ourwhite} + Refiner (Ours) & 2.2B & \phantom{0}\samebf{1.76}\improvement{0.16} & 319.4 & 0.80 & 0.62 \\
        \midrule
        M-VAR-d32~\cite{ren2024m} & 3.0B & \phantom{0}1.78\improvementphantom & 331.2 & 0.83 & 0.61 \\
        % \bottomrule
        % Method & Params & FID$\downarrow$ & IS$\uparrow$ & Prec$\uparrow$ & Rec$\uparrow$ \\
        \midrule
        \midrule
        \multicolumn{4}{l}{\color{ourgray} \textbf{\textit{Generative Adversarial Nets~\cite{goodfellow2014generative}}}} \\
        \color{ourgray} BigGAN-deep~\cite{brock2018large} & \color{ourgray} 112M & \color{ourgray} \phantom{0}6.95\improvementphantom & \color{ourgray} 202.6 & \color{ourgray} 0.87 & \color{ourgray} 0.28 \\
        % GigaGAN~\cite{kang2023scaling} & \color{ourgray} 569M & \color{ourgray} \phantom{0}3.45 & \color{ourgray} 225.5 \\
        \color{ourgray} StyleGAN-XL~\cite{sauer2022stylegan} & \color{ourgray} 166M & \color{ourgray} \phantom{0}2.30\improvementphantom & \color{ourgray} 265.1 & \color{ourgray} 0.78 & \color{ourgray} 0.53 \\
        \multicolumn{4}{l}{\textbf{\color{ourgray} \textit{Diffusion}~\cite{song2020score,ho2020denoising}}} \\
        % \multicolumn{4}{l}{\TODO{}} \\
        \color{ourgray} ADM-G~\cite{dhariwal2021diffusion} & \color{ourgray} 554M & \color{ourgray}  \phantom{0}4.59\improvementphantom & \color{ourgray} 186.7 & \color{ourgray} 0.82 & \color{ourgray} 0.52 \\
        \color{ourgray} LDM-4-G~\cite{rombach2022high} & \color{ourgray} 400M & \color{ourgray} \phantom{0}3.60\improvementphantom & \color{ourgray} 247.6 & \color{ourgray} -- & \color{ourgray} -- \\
        \color{ourgray} DiT-XL/2~\cite{peebles2023scalable} & \color{ourgray} 675M & \color{ourgray} \phantom{0}2.27\improvementphantom & \color{ourgray} 278.2 & \color{ourgray} 0.83 & \color{ourgray} 0.57 \\
        \color{ourgray} SiT-XL/2~\cite{ma2024sit} & \color{ourgray} 675M & \color{ourgray} \phantom{0}2.06\improvementphantom & \color{ourgray} 252.2 & \color{ourgray} -- & \color{ourgray} -- \\
        \color{ourgray} JiT-G/16~\cite{li2025back} & \color{ourgray} 2.0B & \color{ourgray} \phantom{0}1.82\improvementphantom & \color{ourgray} 292.6 & \color{ourgray} 0.79 & \color{ourgray} 0.62 \\
        \color{ourgray} FlowAR~\cite{ren2024flowar} & \color{ourgray} 1.9B & \color{ourgray} \phantom{0}1.65\improvementphantom & \color{ourgray} 296.5 & \color{ourgray} 0.83 & \color{ourgray} 0.60 \\
        % \TODO{} \\
        \multicolumn{4}{l}{\textbf{\color{ourgray} \textit{Raster Autoregression}}} \\
        \color{ourgray} VQGAN~\cite{esser2021taming} & \color{ourgray} 1.4B & \color{ourgray} 15.78\improvementphantom & \color{ourgray} \phantom{0}74.3 & \color{ourgray} -- & \color{ourgray} -- \\
        \color{ourgray} RQ-Transformer~\cite{lee2022draft} & \color{ourgray} 3.8B & \color{ourgray} \phantom{0}7.55\improvementphantom & \color{ourgray} 134.0 & \color{ourgray} 0.73 & \color{ourgray} 0.58 \\
        \color{ourgray} LlamaGen-3B~\cite{sun2024autoregressive} & \color{ourgray} 3.1B & \color{ourgray} \phantom{0}2.18\improvementphantom & \color{ourgray} 263.3 & \color{ourgray} 0.81 & \color{ourgray} 0.58 \\
        \color{ourgray} RAR-XXL~\cite{yu2025randomized} & \color{ourgray} 1.5B & \color{ourgray} \phantom{0}1.48\improvementphantom & \color{ourgray} 326.0 & \color{ourgray} 0.80 & \color{ourgray} 0.63 \\
        % \multicolumn{4}{l}{\TODO{}} \\
        \multicolumn{4}{l}{\textbf{\color{ourgray} \textit{Masked Autoregression}}} \\
        \color{ourgray} MAGE~\cite{li2023mage} & \color{ourgray} 439M & \color{ourgray} \phantom{0}7.04\improvementphantom & \color{ourgray} 123.5 & \color{ourgray} -- & \color{ourgray} -- \\
        \color{ourgray} MaskGiT~\cite{chang2022maskgit} & \color{ourgray} 227M & \color{ourgray} \phantom{0}6.18\improvementphantom & \color{ourgray} 182.1 & \color{ourgray} 0.80 & \color{ourgray} 0.51 \\
        \color{ourgray} MAR-H~\cite{li2024autoregressive} & \color{ourgray} 943M & \color{ourgray} \phantom{0}1.55\improvementphantom & \color{ourgray} 303.7 & \color{ourgray} 0.81 & \color{ourgray} 0.62 \\
        \midrule
        % \color{ourgray} ImageNet Validation & \color{ourgray} -- & \color{ourgray} 1.78\improvementphantom & \color{ourgray} -- & \color{ourgray} -- & \color{ourgray} -- \\
        ImageNet Validation & -- & 1.78\improvementphantom & -- & -- & -- \\
        \bottomrule
    \end{tabular}
    }}
\end{table}

\begin{table}[t]
    \centering
    \caption{\textbf{Extra Evaluations without CFG across Scales.} Similar to the typical inference setting with CFG \textbf{(a)}, our refiner also enables significant gains in generative quality without it \textbf{(b)}.}
    \newcommand{\improvement}[1]{$_{\color{ourgreen}\blacktriangledown #1}$}
    \newcommand{\worsened}[1]{$_{\color{ourred}\blacktriangle #1}$}
    \newcommand{\improvementphantom}{\phantom{\improvement{0.00}}}
    \adjustbox{max width=.8\linewidth}{
        % \begin{tabular}{l c@{}c@{}c@{}c c@{}c@{}c@{}c}
        \begin{tabular}{l cccc c cccc}
            \toprule
            \multirow{2}{*}[-.2em]{FID ($\downarrow$)} & \multicolumn{4}{c}{\textbf{(a)} CFG: \cmark{}} & & \multicolumn{4}{c}{\textbf{(b)} CFG: \xmark{}} \\
            \cmidrule{2-5}\cmidrule{7-10}
            & d16 & d20 & d24 & d30 & & d16 & d20 & d24 & d30 \\
            \midrule
            VAR & 3.30\improvementphantom & 2.57\improvementphantom & 2.09\improvementphantom & 1.92\improvementphantom & & 3.44\improvementphantom  & 2.62\improvementphantom  & 2.13\improvementphantom  & 2.17\improvementphantom \\
            + Refiner (Ours) & 2.81\improvement{0.49} & 2.17\improvement{0.40} & 1.83\improvement{0.26} & 1.76\improvement{0.16} & & {3.41}\improvement{0.03} & {2.40}\improvement{0.22} & 1.94\improvement{0.19} & 1.88\improvement{0.29} \\
            \bottomrule
        \end{tabular}
    }
    \label{tab:unguided}
\end{table}

\section{Additional Qualitative Samples}

\subsection{Text-to-Image}\label{sec:app_qualitative_t2i}
\begin{figure*}[t]
  \centering
  \adjustbox{max height=.7\linewidth}{
  \makebox[\textwidth]{%
    \begin{minipage}{0.80\textwidth}
      \centering

      % Column headers
      \begin{minipage}[t]{0.48\linewidth}
        \centering\textbf{Baseline}
      \end{minipage}
      \hfill
      \begin{minipage}[t]{0.48\linewidth}
        \centering\textbf{Ours}
      \end{minipage}

      \vspace{4pt}

      % Row 1
      \begin{minipage}[t]{0.48\linewidth}
        \centering
        \includegraphics[width=\linewidth]{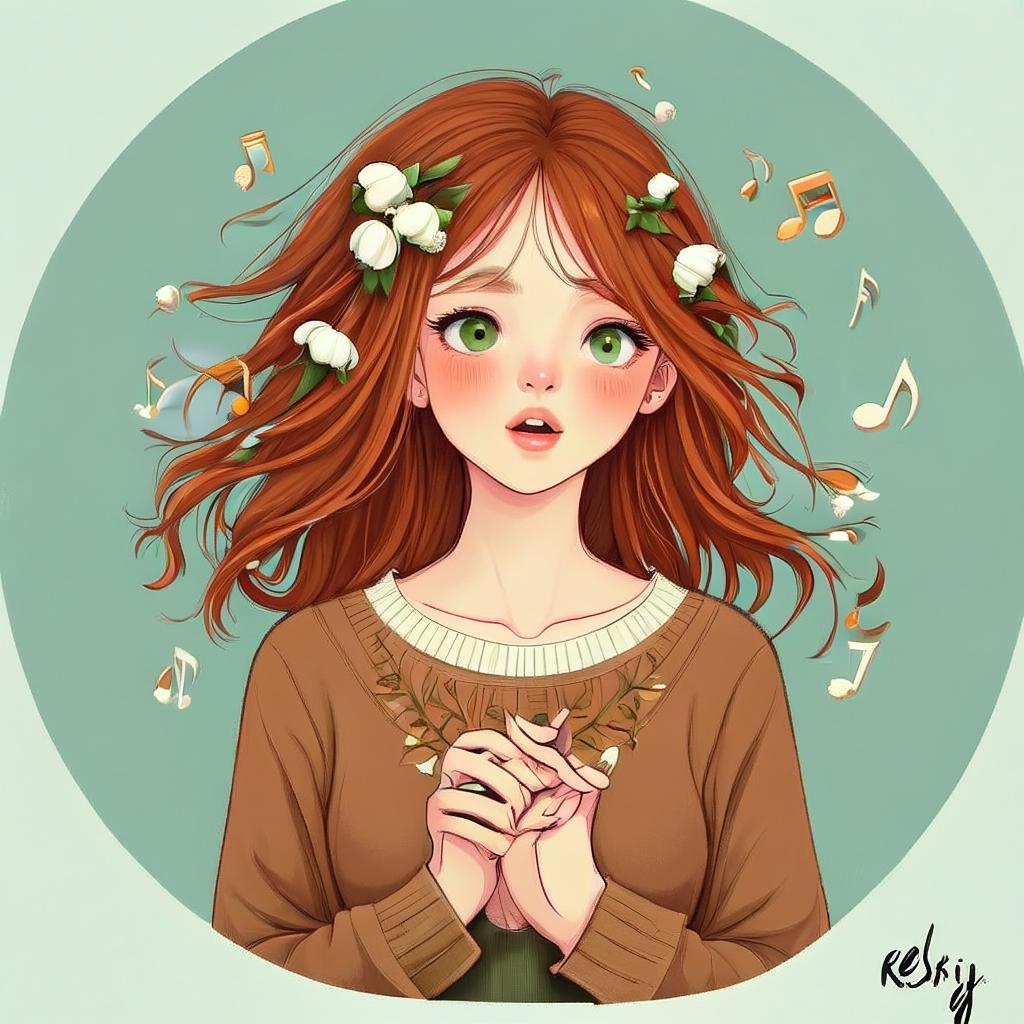}
      \end{minipage}
      \hfill
      \begin{minipage}[t]{0.48\linewidth}
        \centering
        \includegraphics[width=\linewidth]{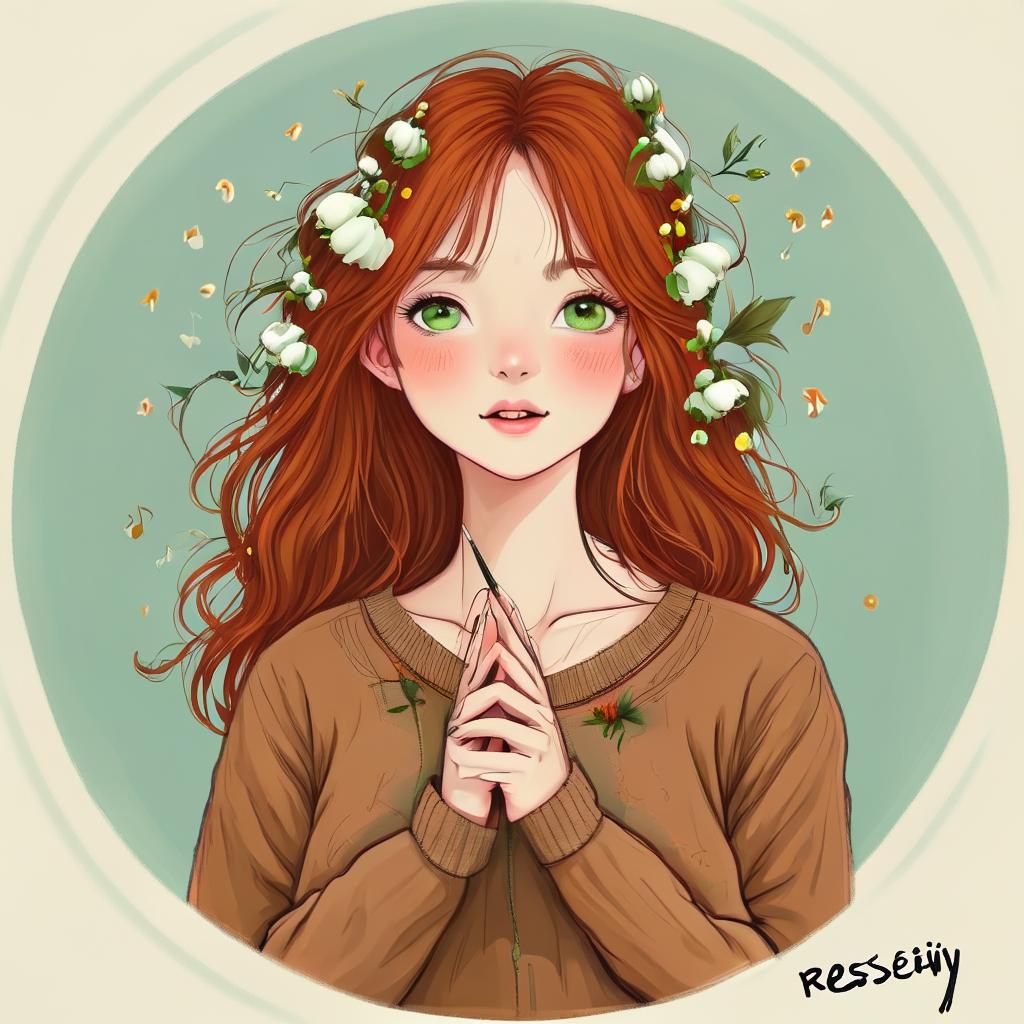}
      \end{minipage}

        \vspace{2pt}
        \begin{center}
          \parbox{0.9\linewidth}{%
            \centering\footnotesize\linespread{0.9}\selectfont
The image showcases a detailed and enchanting illustration of a young woman, possibly a sketch in progress, set against a light teal circular backdrop ...
          }
        \end{center}

      \vspace{8pt}

      % Row 2
      \begin{minipage}[t]{0.48\linewidth}
        \centering
        \includegraphics[width=\linewidth]{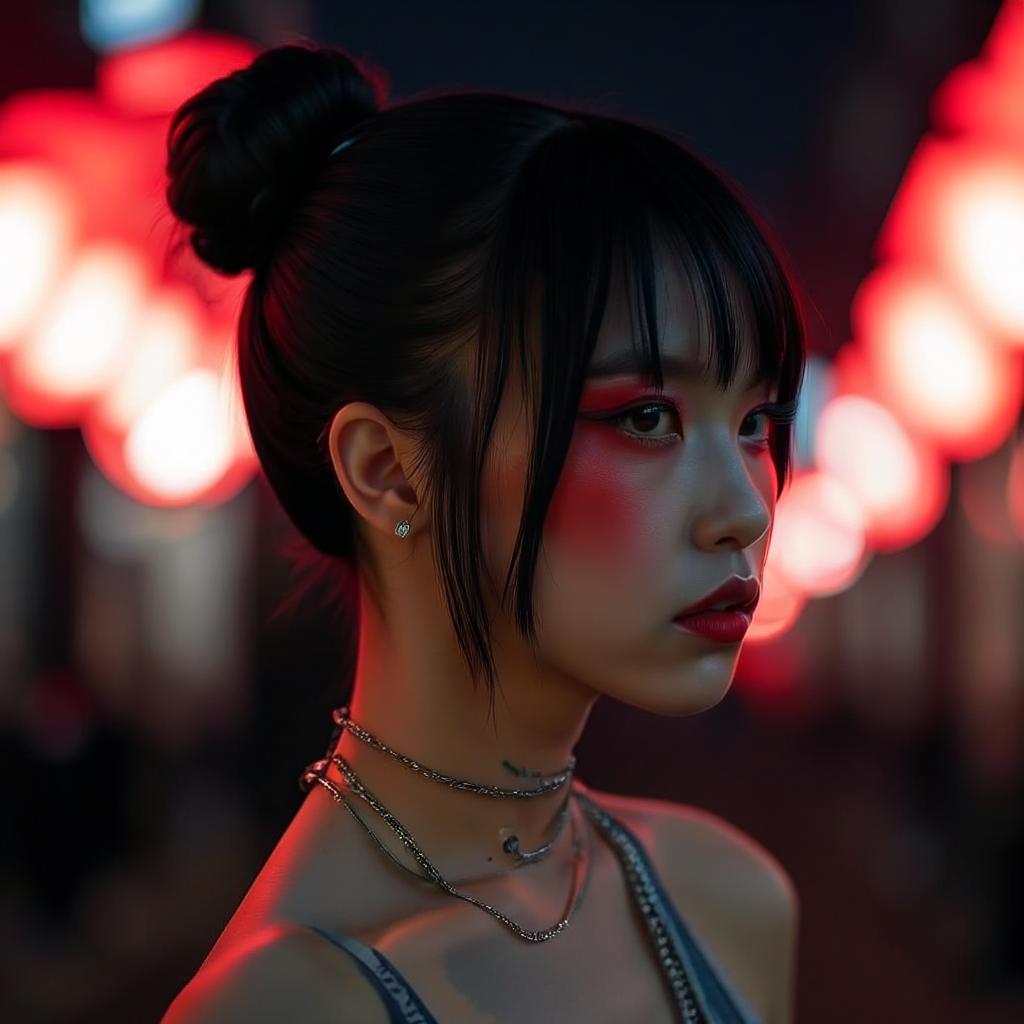}
      \end{minipage}
      \hfill
      \begin{minipage}[t]{0.48\linewidth}
        \centering
        \includegraphics[width=\linewidth]{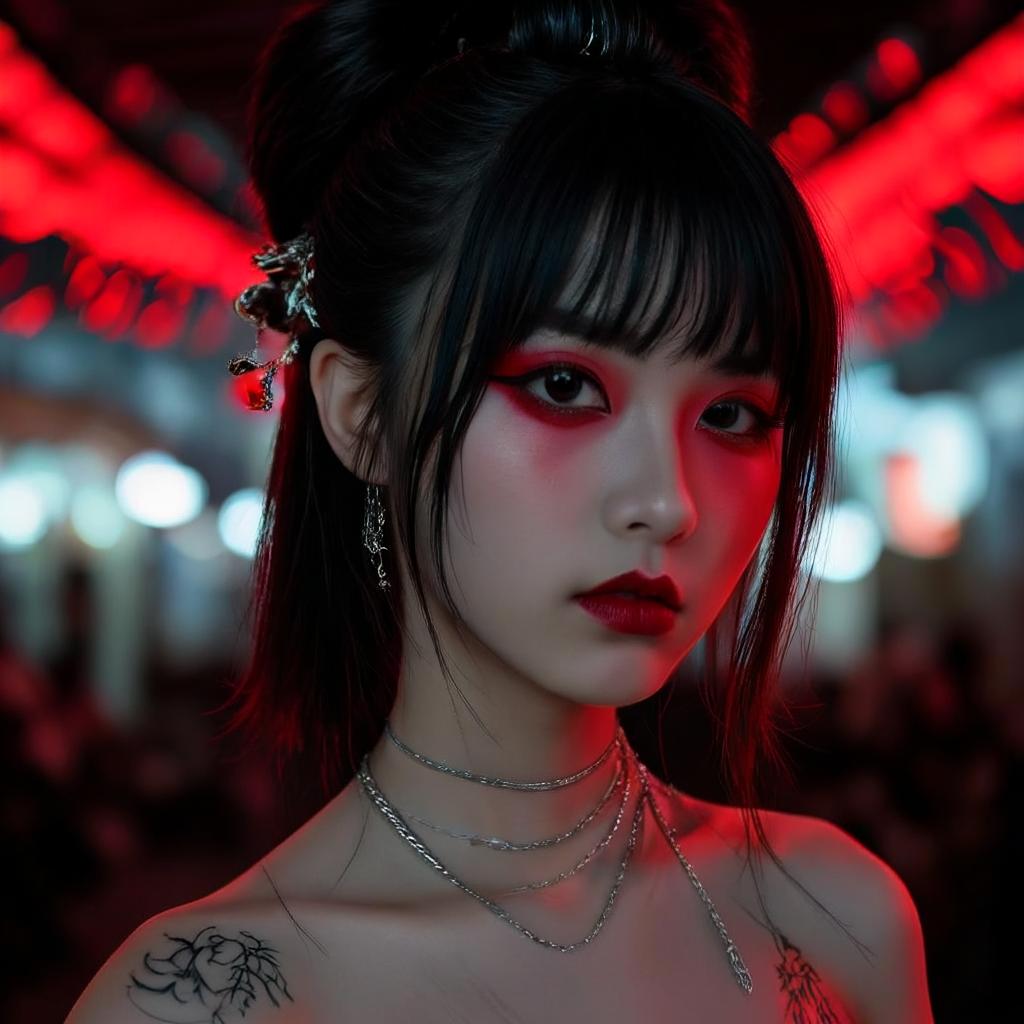}
      \end{minipage}

      \vspace{2pt}
      \small The image features a young woman with a striking blend of modern and traditional Japanese-inspired aesthetics. She is captured from the chest up, with her gaze slightly averted to the right. ...

      \vspace{8pt}

      % Row 3
      \begin{minipage}[t]{0.48\linewidth}
        \centering
        \includegraphics[width=\linewidth]{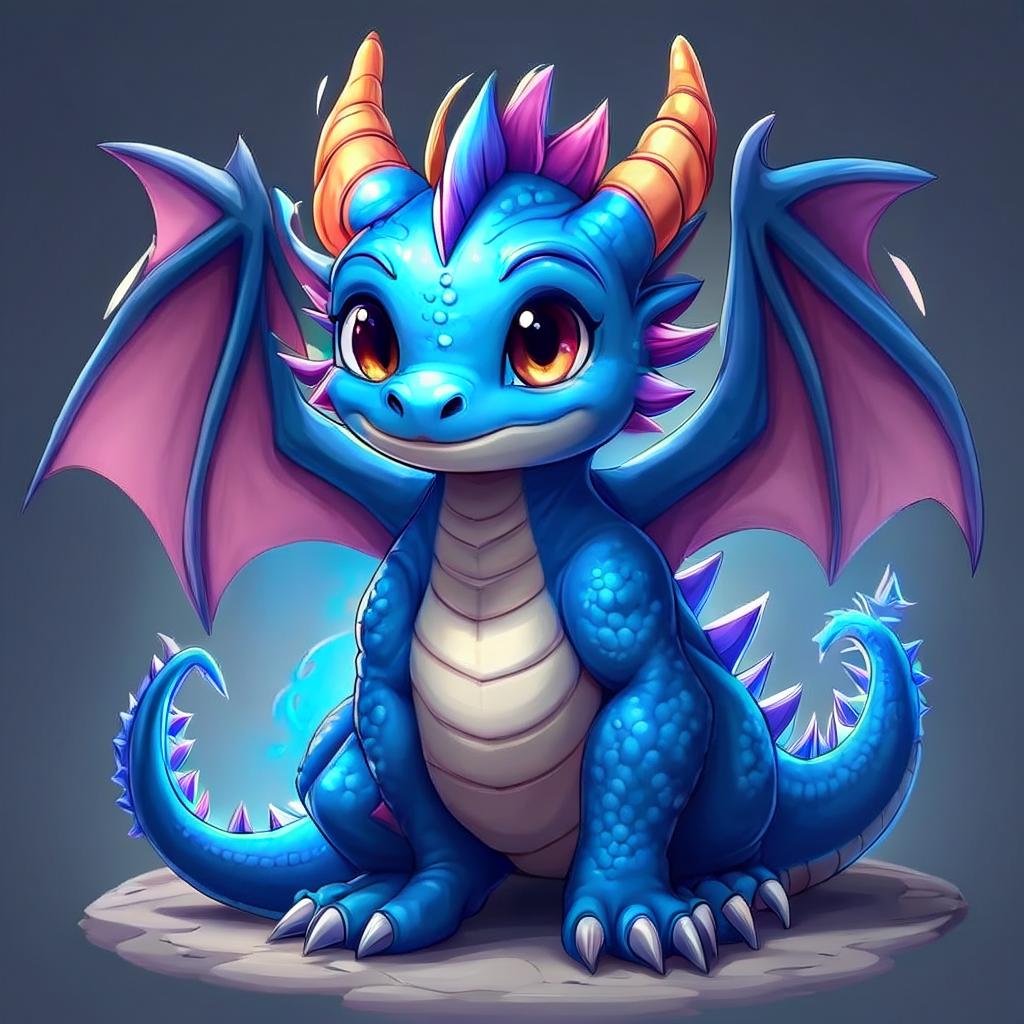}
      \end{minipage}
      \hfill
      \begin{minipage}[t]{0.48\linewidth}
        \centering
        \includegraphics[width=\linewidth]{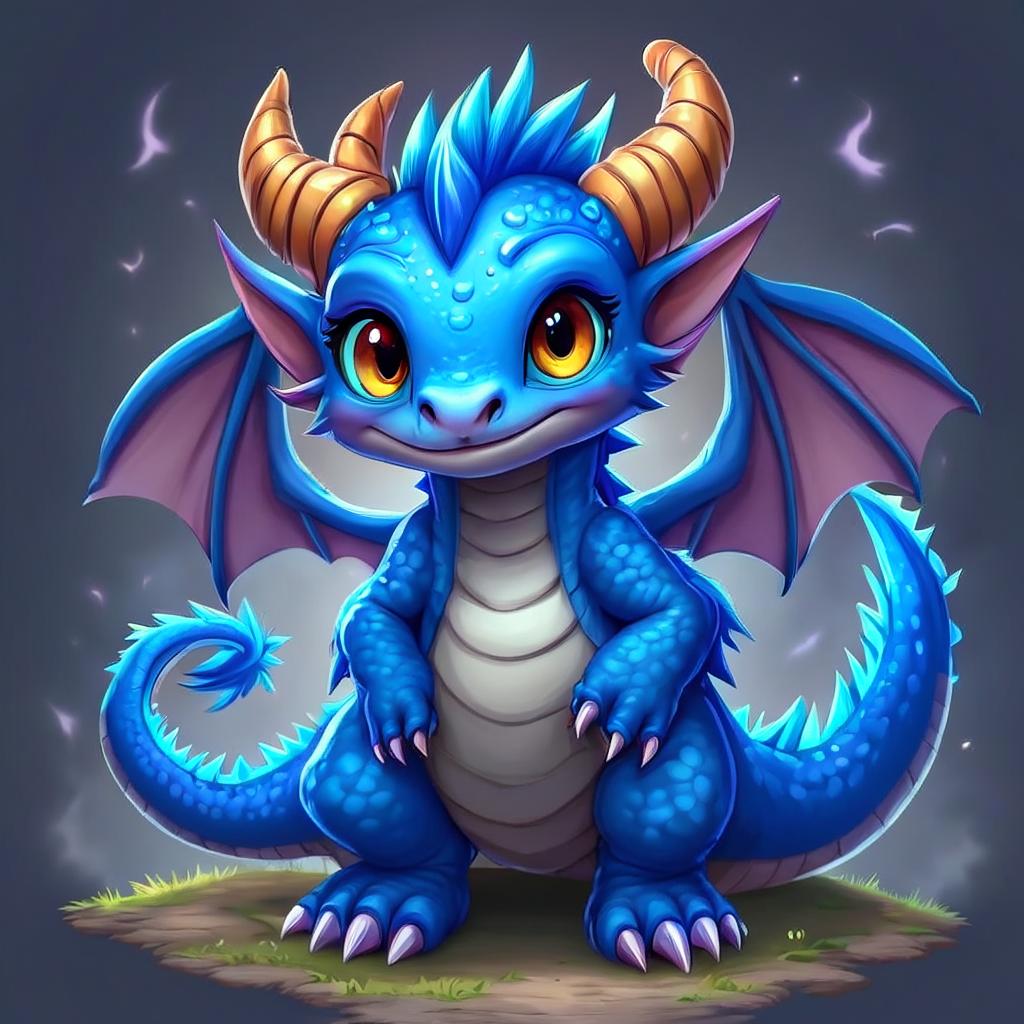}
      \end{minipage}

      \vspace{2pt}
      \small world of warcraft Malygos, the blue dragon aspect, cute tee shirt design illustration, 4k

      \vspace{8pt}
    \end{minipage}%
  }
  }
  \caption{Selected Infinity-2B + refiner samples on HPSv3~\cite{ma2025hpsv3} prompts.}
  \label{fig:app_t2i_selected_1}
\end{figure*}

\begin{figure*}[t]
  \centering
  \adjustbox{max height=.7\linewidth}{
  \makebox[\textwidth]{%
    \begin{minipage}{0.80\textwidth}
      \centering

      % Column headers
      \begin{minipage}[t]{0.48\linewidth}
        \centering\textbf{Baseline}
      \end{minipage}
      \hfill
      \begin{minipage}[t]{0.48\linewidth}
        \centering\textbf{Ours}
      \end{minipage}

      \vspace{4pt}

      % Row 1
      \begin{minipage}[t]{0.48\linewidth}
        \centering
        \includegraphics[width=\linewidth]{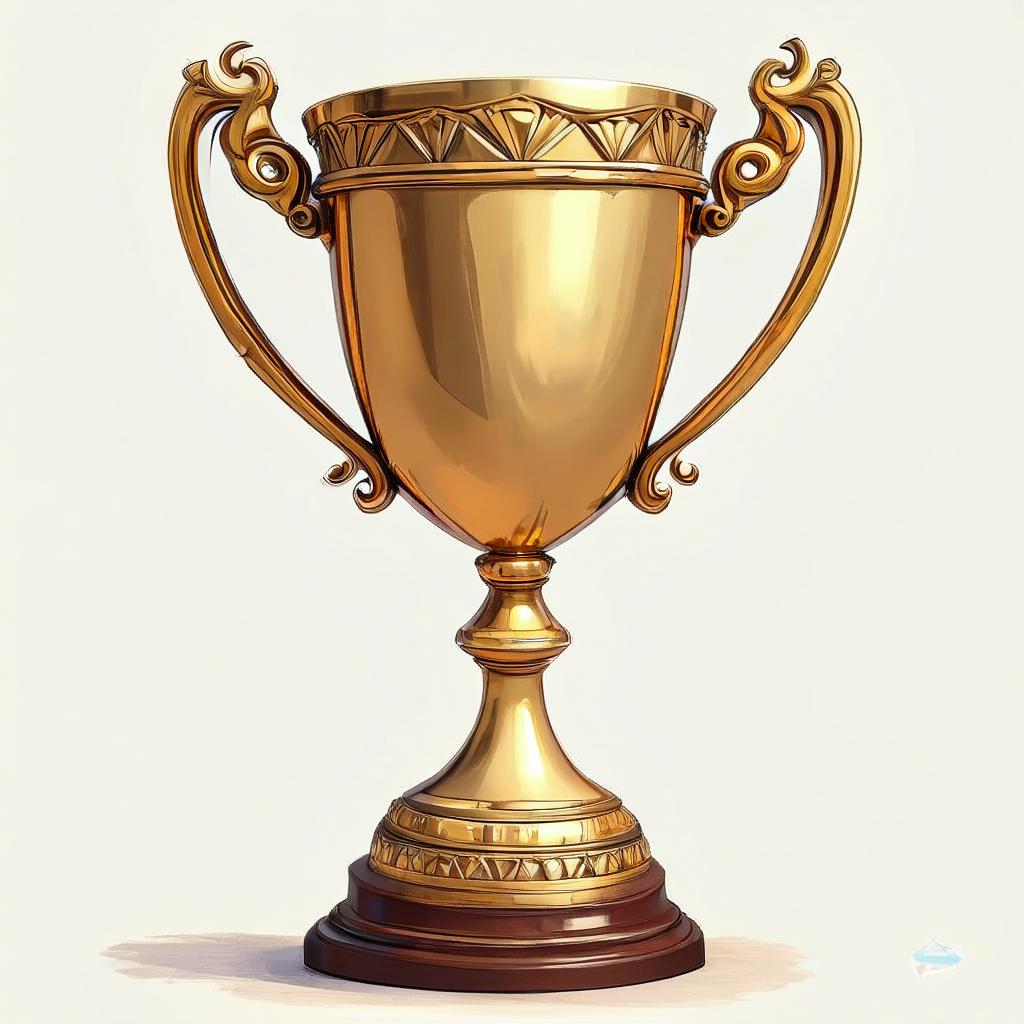}
      \end{minipage}
      \hfill
      \begin{minipage}[t]{0.48\linewidth}
        \centering
        \includegraphics[width=\linewidth]{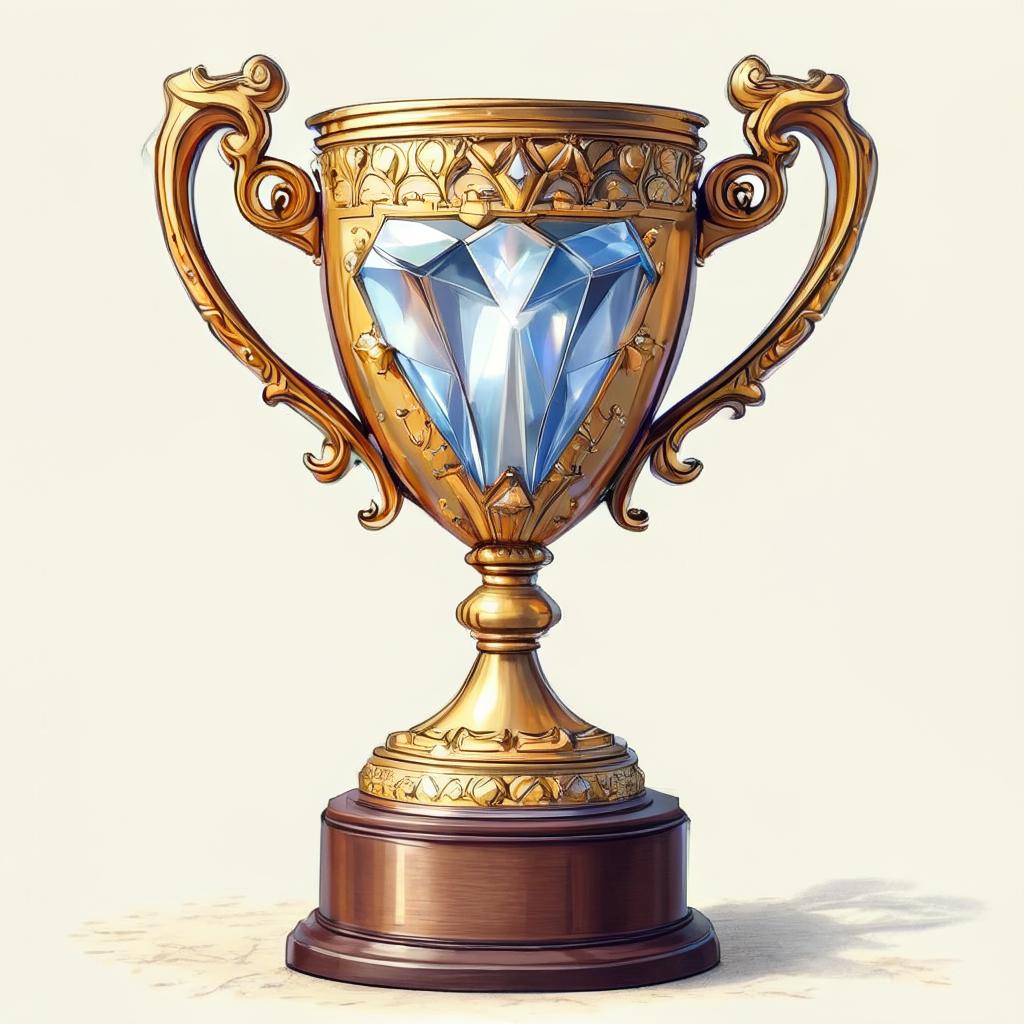}
      \end{minipage}

      \vspace{2pt}
      \small diamond trophy, 2d drawing

      \vspace{8pt}

      % Row 2
      \begin{minipage}[t]{0.48\linewidth}
        \centering
        \includegraphics[width=\linewidth]{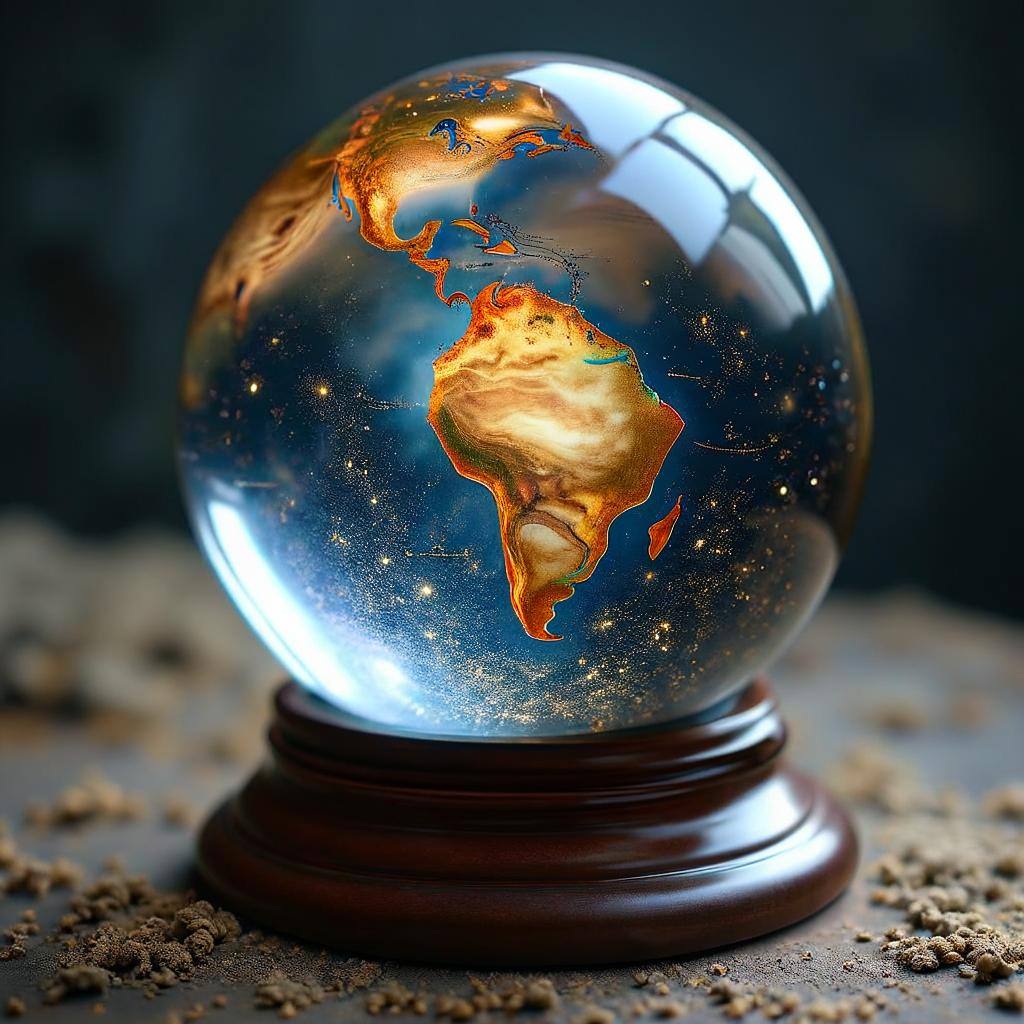}
      \end{minipage}
      \hfill
      \begin{minipage}[t]{0.48\linewidth}
        \centering
        \includegraphics[width=\linewidth]{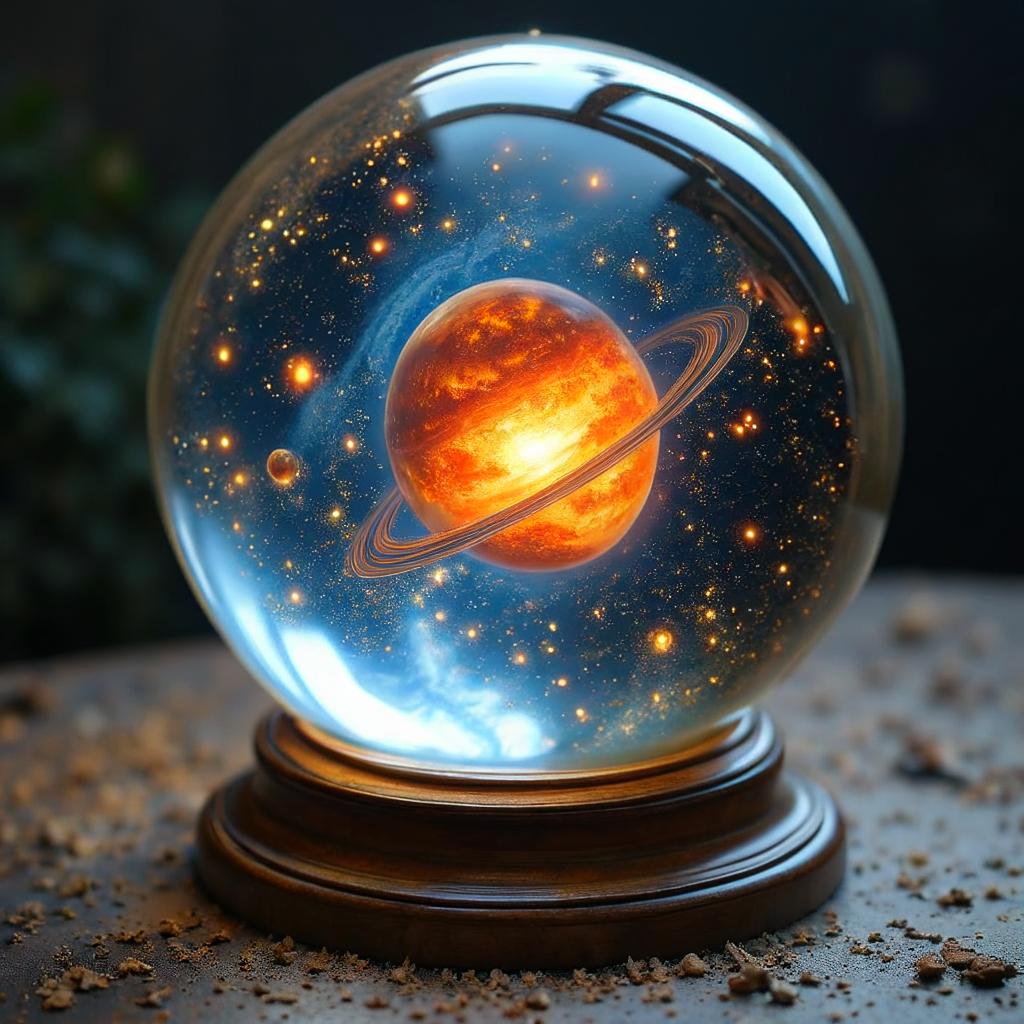}
      \end{minipage}

      \vspace{2pt}
      \small The whole universe enclosed in a glass globe, exquisite detail.

      \vspace{8pt}

      % Row 3
      \begin{minipage}[t]{0.48\linewidth}
        \centering
        \includegraphics[width=\linewidth]{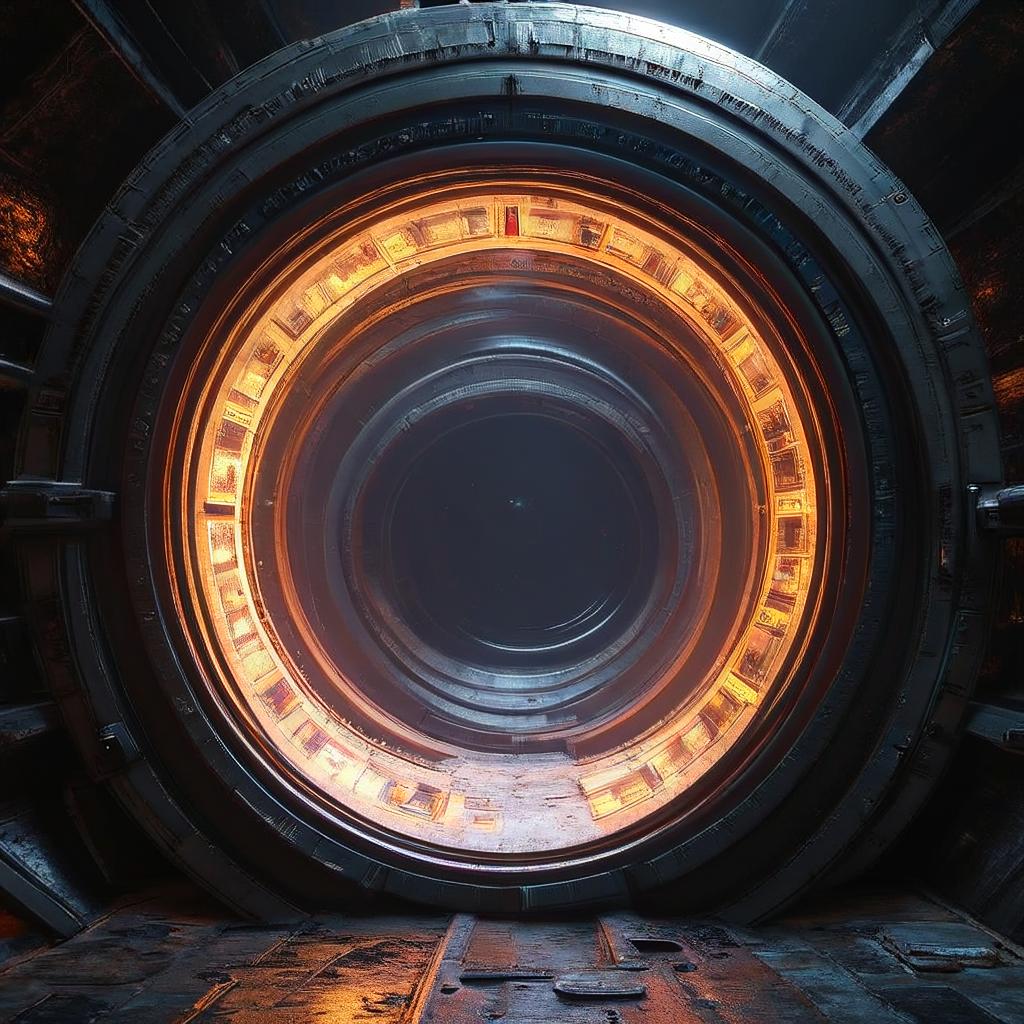}
      \end{minipage}
      \hfill
      \begin{minipage}[t]{0.48\linewidth}
        \centering
        \includegraphics[width=\linewidth]{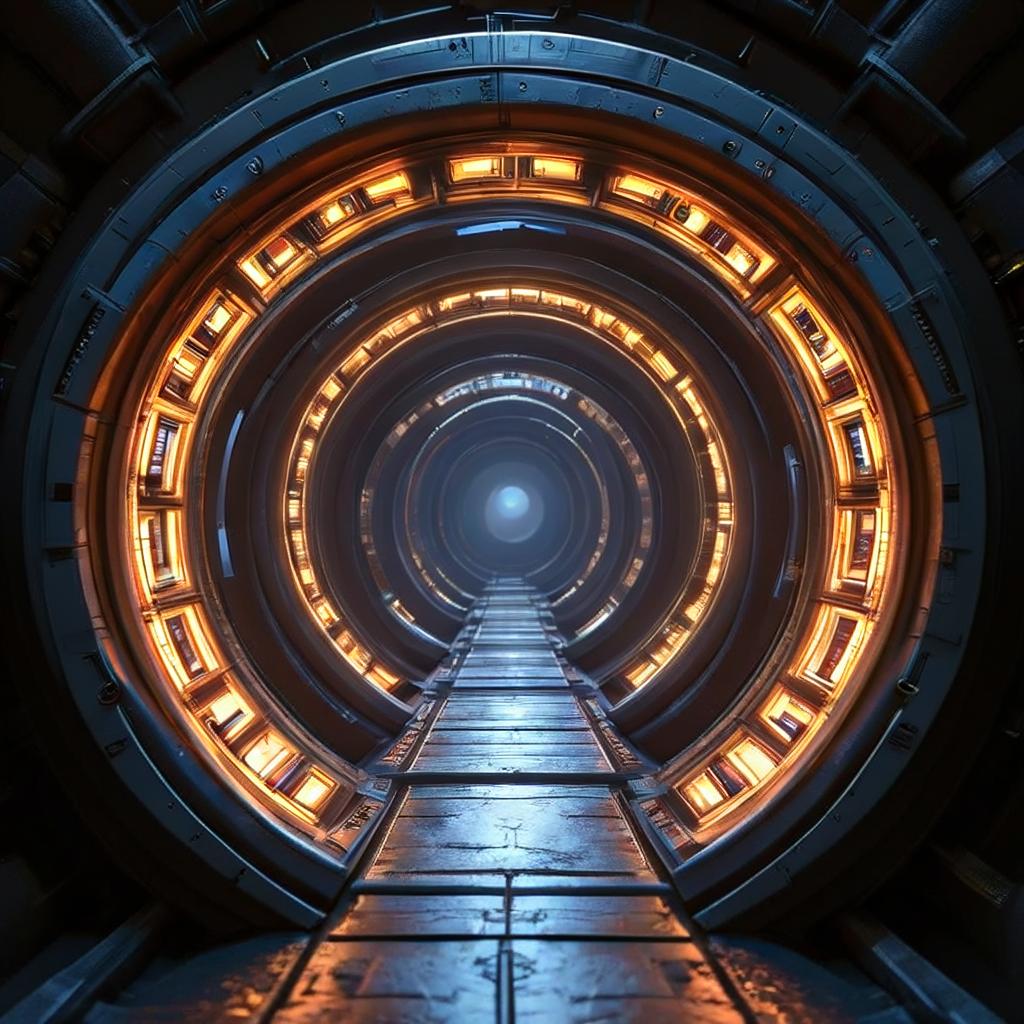}
      \end{minipage}

      \vspace{2pt}
      \small a cylindrical 25th century warp core in the style of "Star Trek"

      \vspace{8pt}
    \end{minipage}%
  }
  }
  \caption{Selected Infinity-2B + refiner samples on HPSv3~\cite{ma2025hpsv3} prompts.}
  \label{fig:app_t2i_selected_2}
\end{figure*}

We show additional comparisons between the base Infinity-2B~\cite{han2025infinity} and our refiner version in \cref{fig:app_t2i_selected_1,fig:app_t2i_selected_2}.

\subsection{ImageNet}\label{sec:app_qualitative_imagenet}
\begin{figure}[t]
    \centering
    \includegraphics[width=\linewidth]{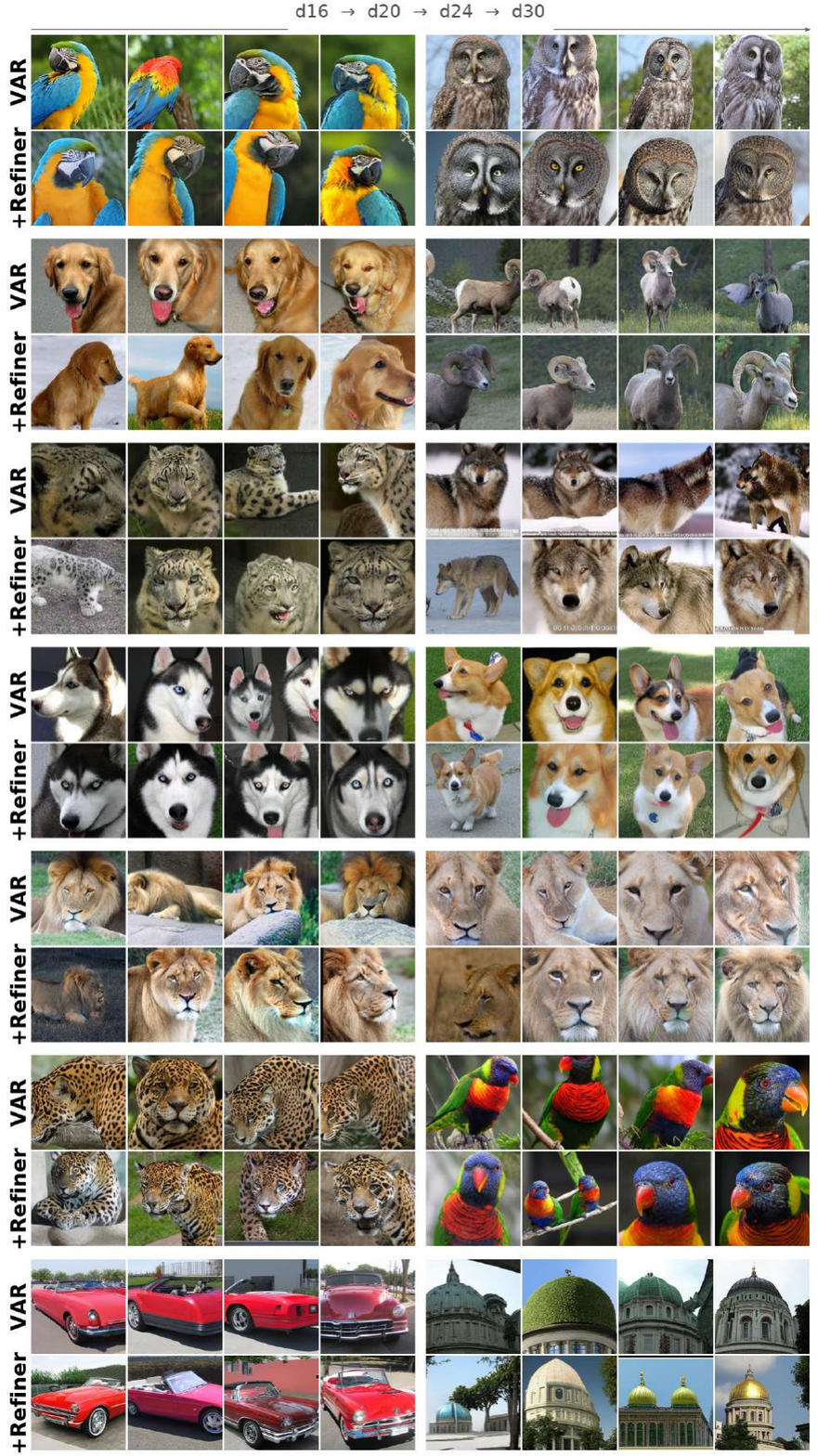}
    \caption{\textbf{Additional Selected Samples from VAR w/o refiner across different scales.} We use a classifier-free guidance scale 2.5 with topk=500 for sampling across the scales.}
    \label{fig:app_baseline_vs_refiner}
\end{figure}

\begin{figure}[t]
    \centering
    
    \begin{subfigure}[t]{0.48\textwidth}
        \centering
        \includegraphics[width=\linewidth]{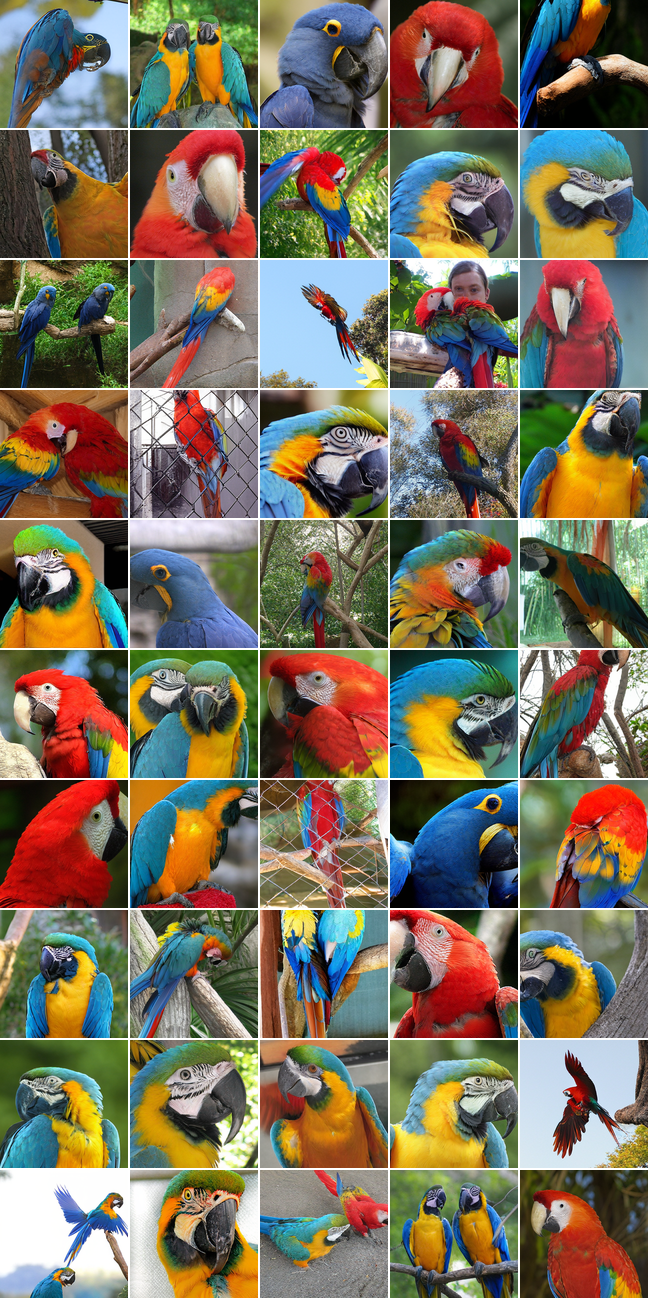}
        \caption{Class: ``macaw'' (88)}
        \label{fig:husky}
    \end{subfigure}
    \hfill
    \begin{subfigure}[t]{0.48\textwidth}
        \centering
        \includegraphics[width=\linewidth]{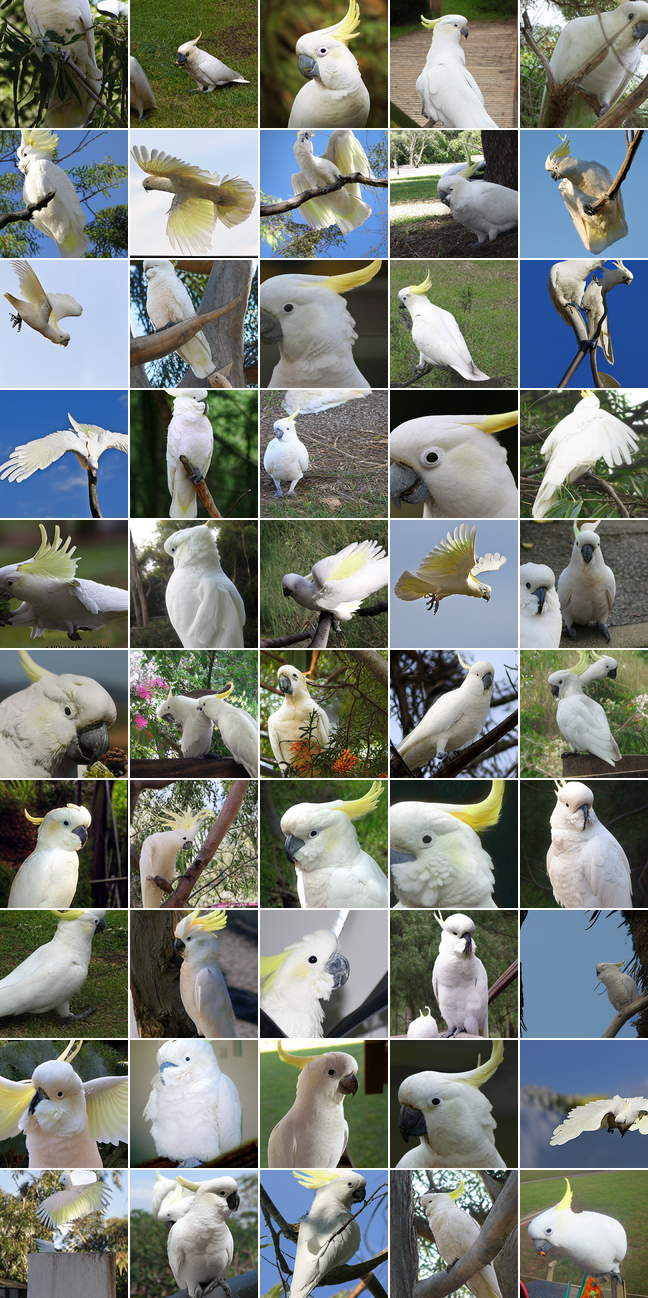}
        \caption{Class: ``Sulphur-crested cockatoo'' (89)}
        \label{fig:cockatoo}
    \end{subfigure}
    \caption{\textbf{Uncurated} 256×256 \textbf{VARd30 + refiner} samples. Classifier-free guidance scale = 2.5, topk = 500.}
    \label{fig:app_imagenet_uncurated_1}
\end{figure}

\begin{figure}[t]
    \centering
    
    \begin{subfigure}{0.48\textwidth}
        \centering
        \includegraphics[width=\linewidth]{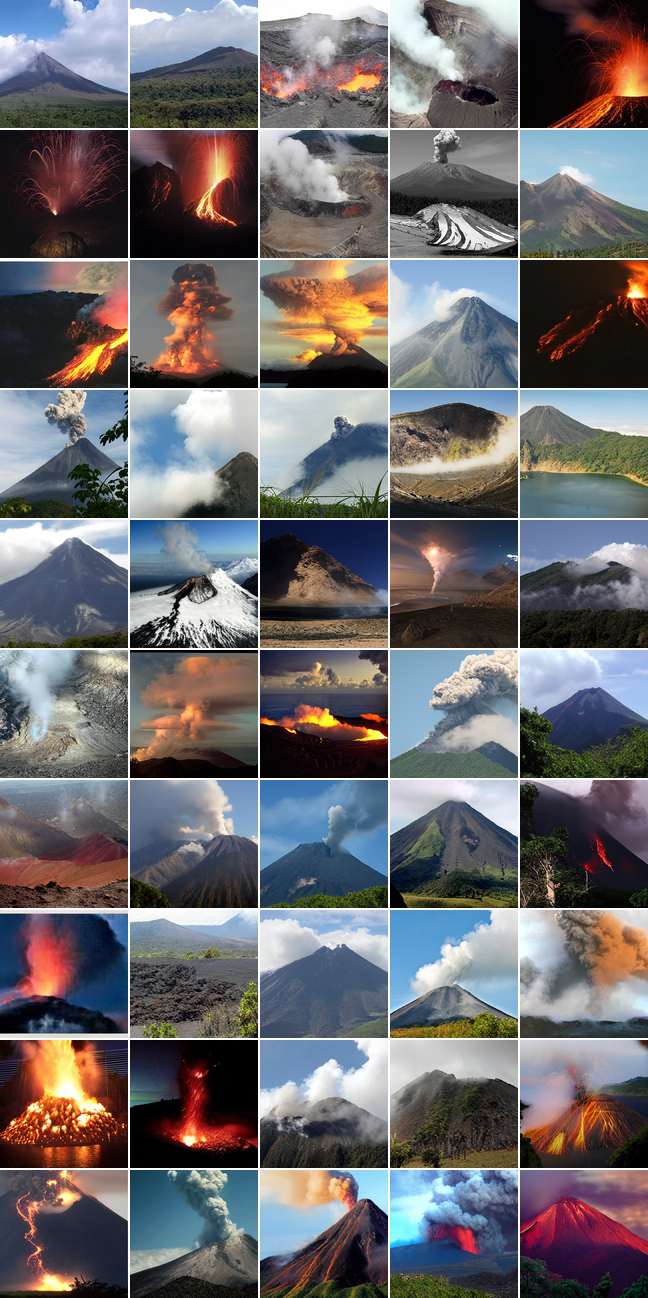}
        \caption{Class: ``alp'' (980)}
        \label{fig:husky}
    \end{subfigure}
    \hfill
    \begin{subfigure}{0.48\textwidth}
        \centering
        \includegraphics[width=\linewidth]{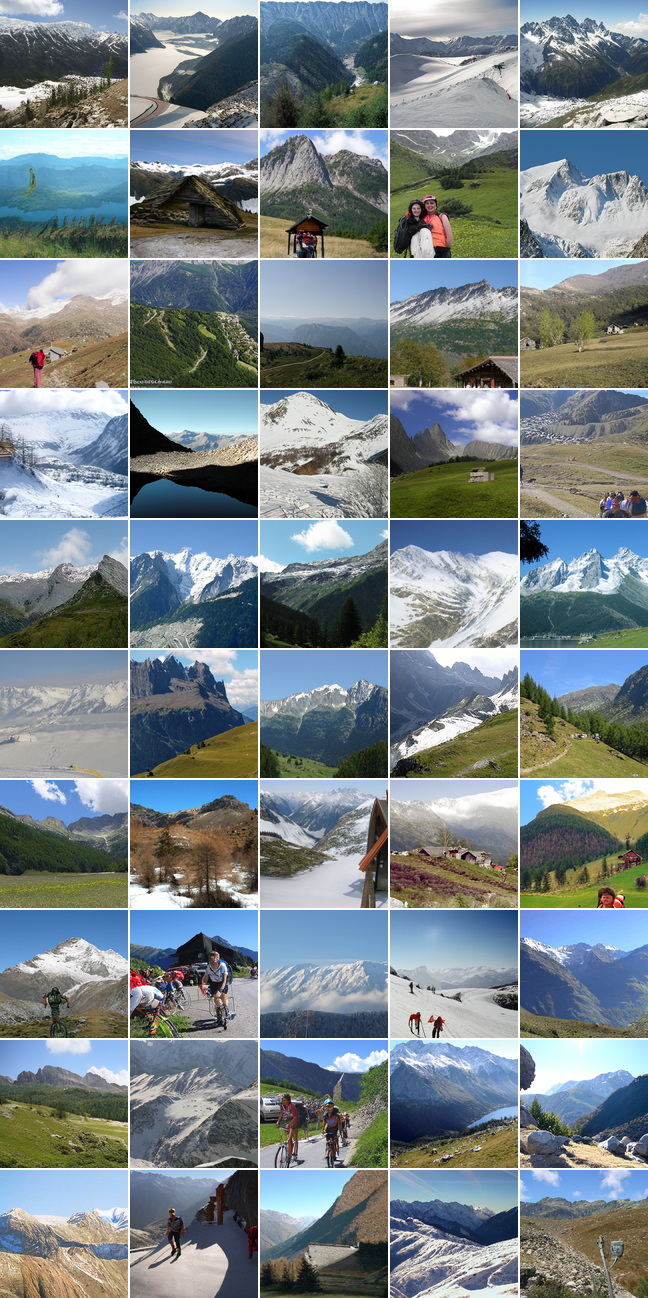}
        \caption{Class: ``volcano'' (970)}
        \label{fig:cockatoo}
    \end{subfigure}
    \caption{\textbf{Uncurated} 256×256 \textbf{VARd30 + refiner} samples. Classifier-free guidance scale = 2.5, topk = 500.}
    \label{fig:app_imagenet_uncurated_2}
\end{figure}

\begin{figure}[t]
    \centering
    
    \begin{subfigure}{0.48\textwidth}
        \centering
        \includegraphics[width=\linewidth]{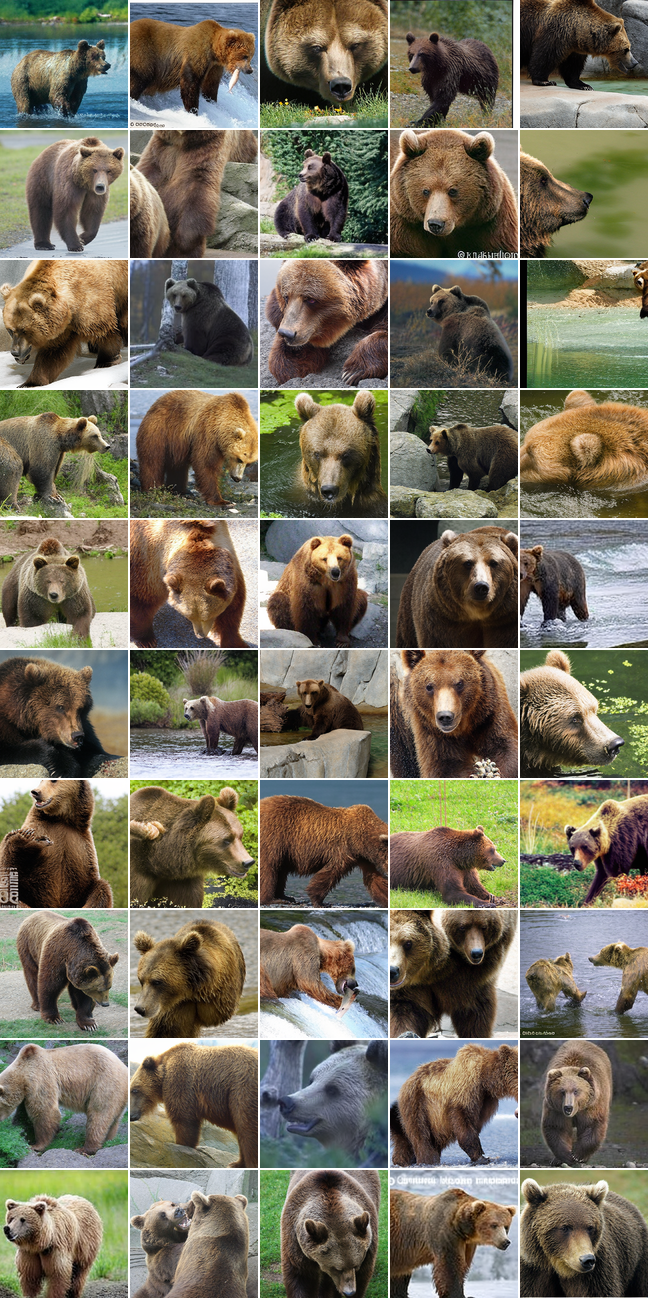}
        \caption{Class: ``brown bear'' (294)}
        \label{fig:husky}
    \end{subfigure}
    \hfill
    \begin{subfigure}{0.48\textwidth}
        \centering
        \includegraphics[width=\linewidth]{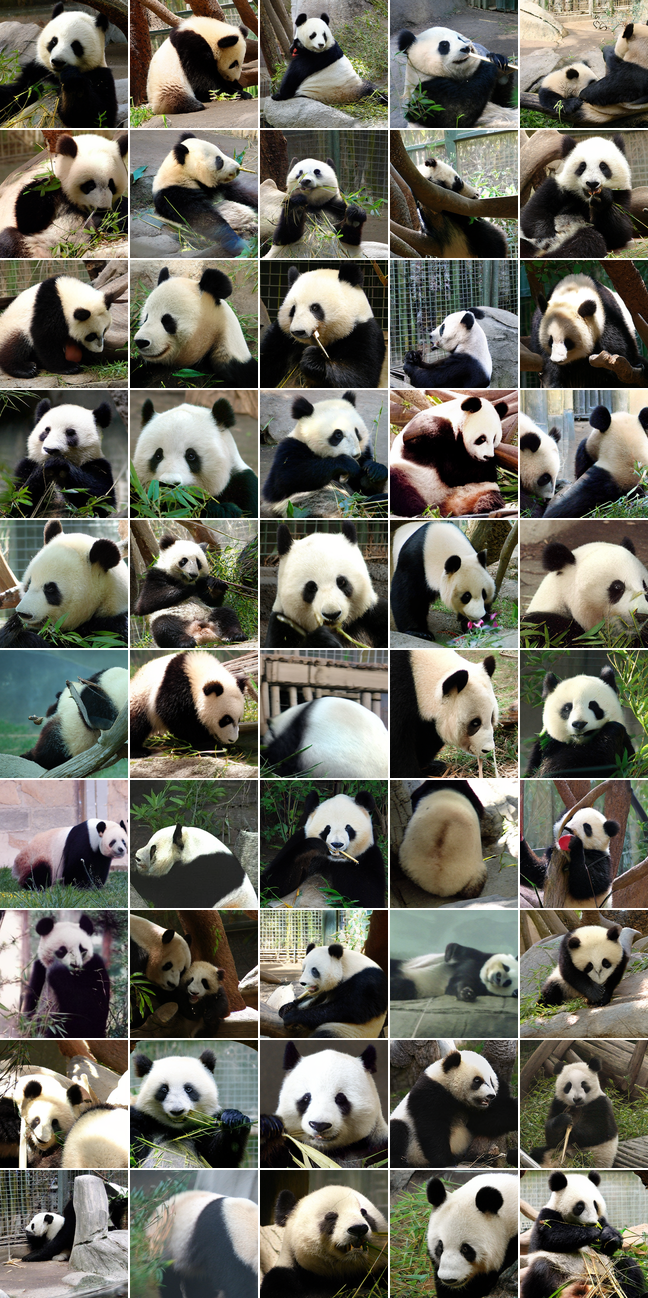}
        \caption{Class: ``giant panda'' (388)}
        \label{fig:cockatoo}
    \end{subfigure}
    \caption{\textbf{Uncurated} 256×256 \textbf{VARd30 + refiner} samples. Classifier-free guidance scale = 2.5, topk = 500.}
    \label{fig:app_imagenet_uncurated_3}
\end{figure}

\begin{figure}[t]
    \centering
    
    \begin{subfigure}{0.48\textwidth}
        \centering
        \includegraphics[width=\linewidth]{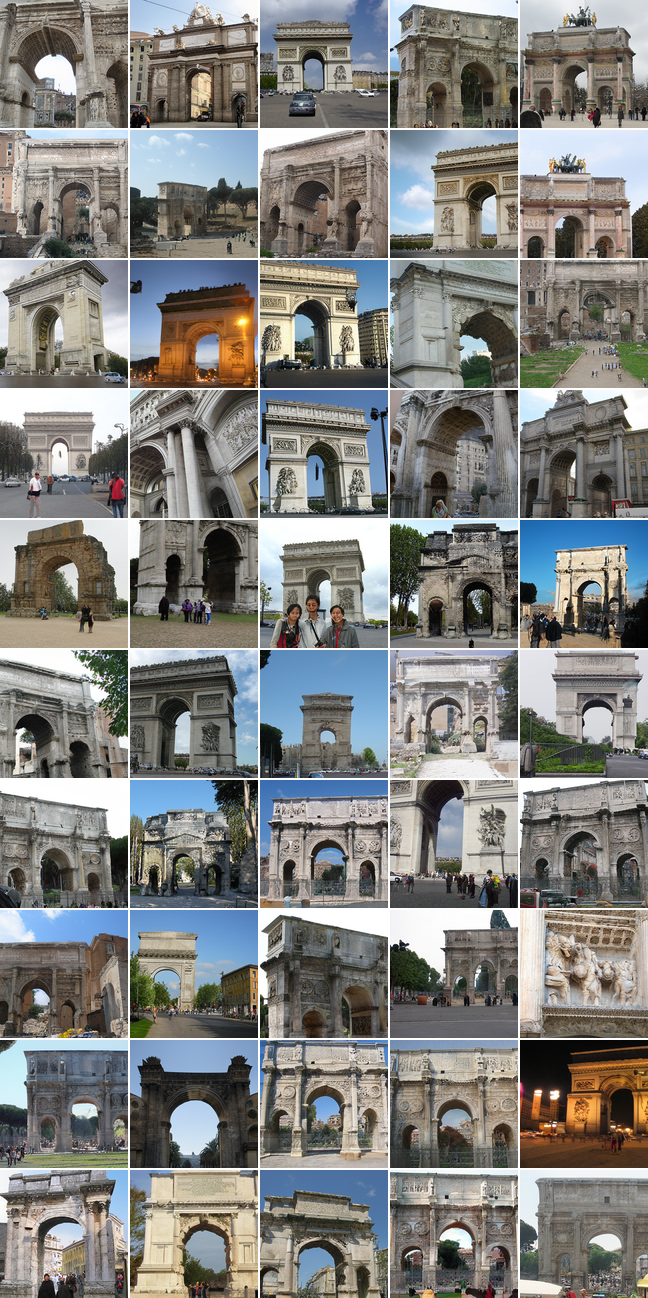}
        \caption{Class: ``triumphal arch'' (873)}
        \label{fig:husky}
    \end{subfigure}
    \hfill
    \begin{subfigure}{0.48\textwidth}
        \centering
        \includegraphics[width=\linewidth]{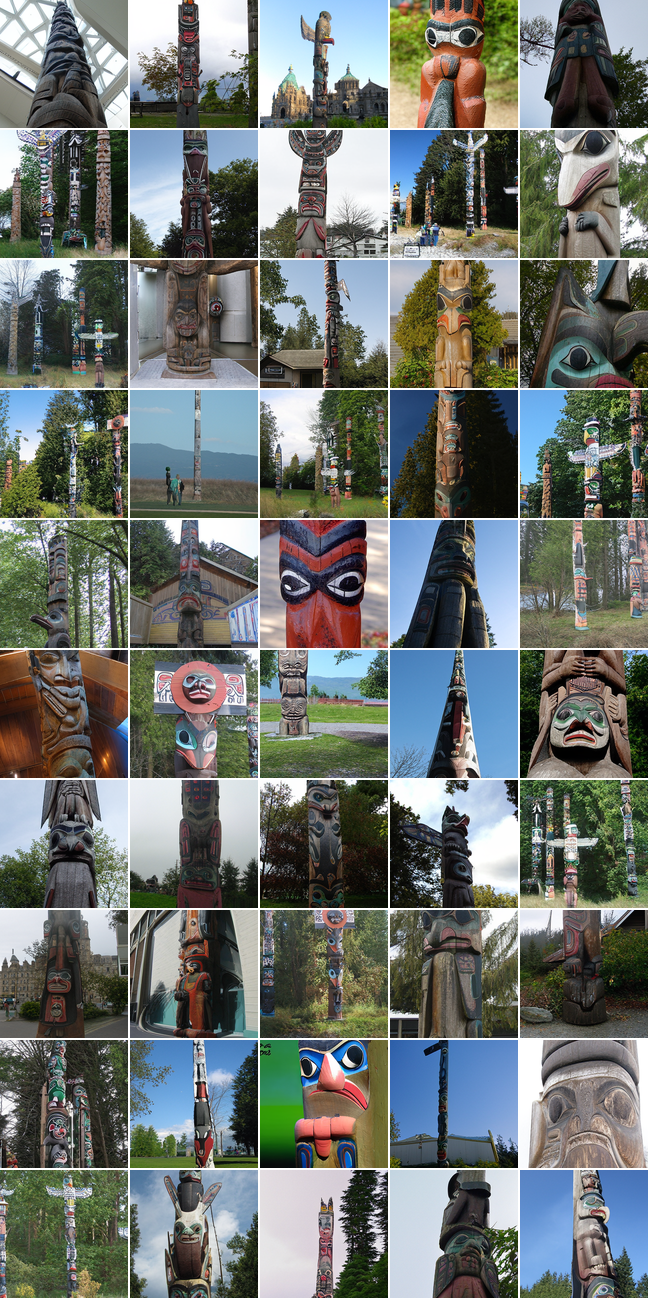}
        \caption{Class: ``totem pole'' (863)}
        \label{fig:cockatoo}
    \end{subfigure}
    \caption{\textbf{Uncurated} 256×256 \textbf{VARd30 + refiner} samples. Classifier-free guidance scale = 2.5, topk = 500.}
    \label{fig:app_imagenet_uncurated_4}
\end{figure}

\begin{figure}[t]
    \centering
    
    \begin{subfigure}{0.48\textwidth}
        \centering
        \includegraphics[width=\linewidth]{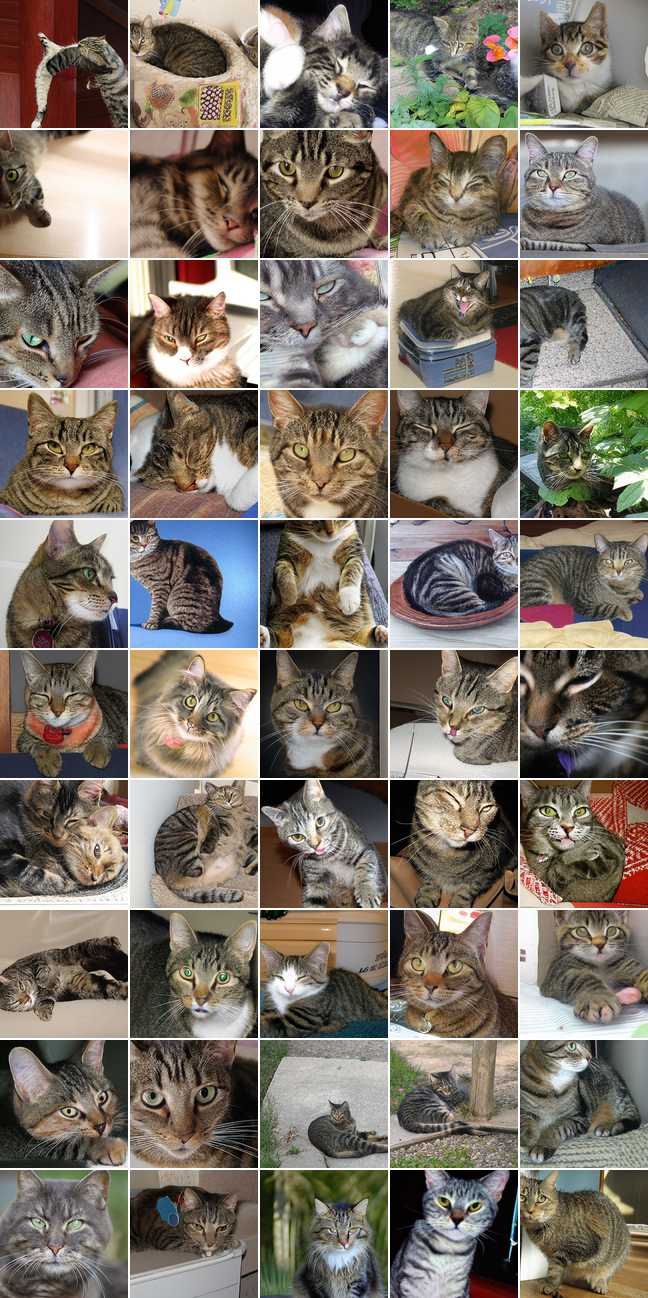}
        \caption{Class: ``tabby'' (281)}
        \label{fig:husky}
    \end{subfigure}
    \hfill
    \begin{subfigure}{0.48\textwidth}
        \centering
        \includegraphics[width=\linewidth]{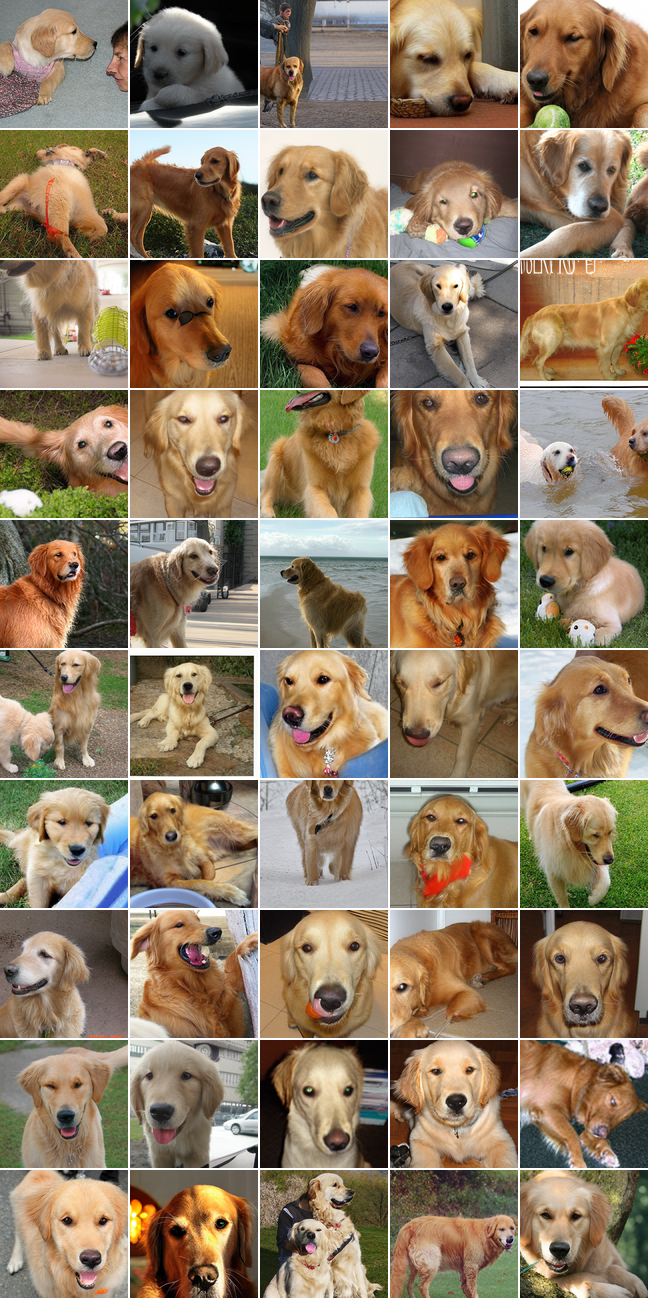}
        \caption{Class: ``golden retriever'' (207)}
        \label{fig:cockatoo}
    \end{subfigure}
    \caption{\textbf{Uncurated} 256×256 \textbf{VARd30 + refiner} samples. Classifier-free guidance scale = 2.5, topk = 500.}
    \label{fig:app_imagenet_uncurated_5}
\end{figure}

\begin{figure}[t]
    \centering
    
    \begin{subfigure}{0.48\textwidth}
        \centering
        \includegraphics[width=\linewidth]{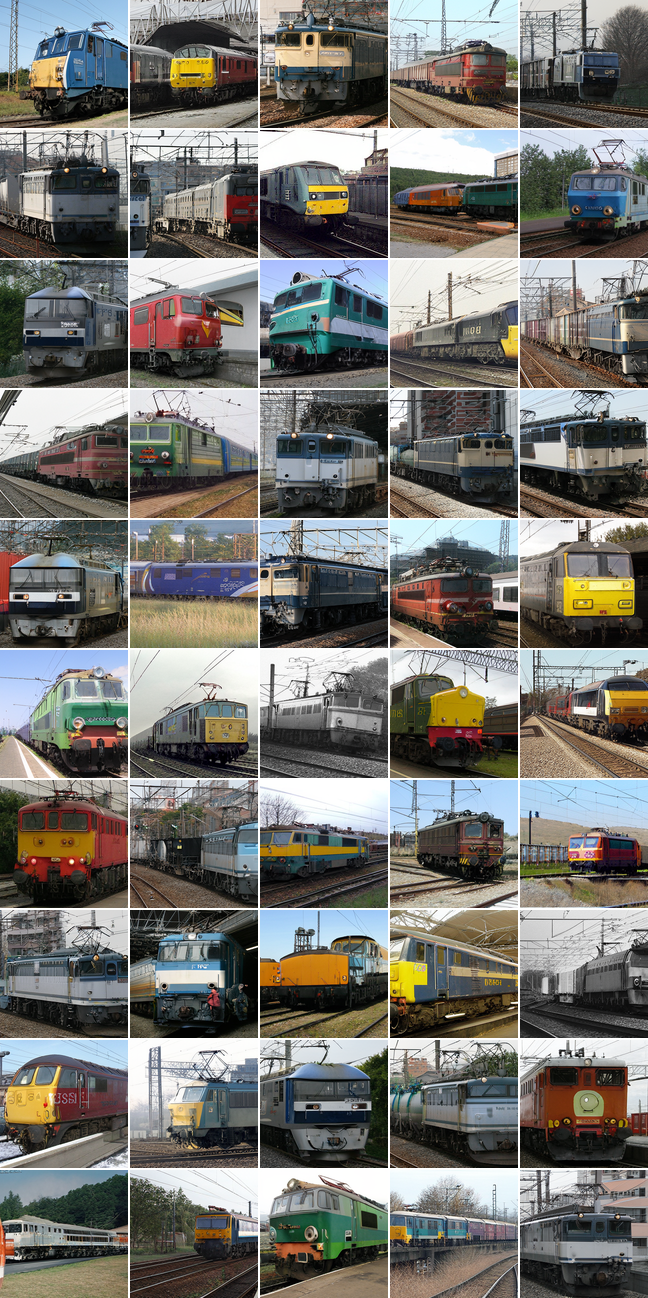}
        \caption{Class: ``electric locomotive'' (547)}
        \label{fig:husky}
    \end{subfigure}
    \hfill
    \begin{subfigure}{0.48\textwidth}
        \centering
        \includegraphics[width=\linewidth]{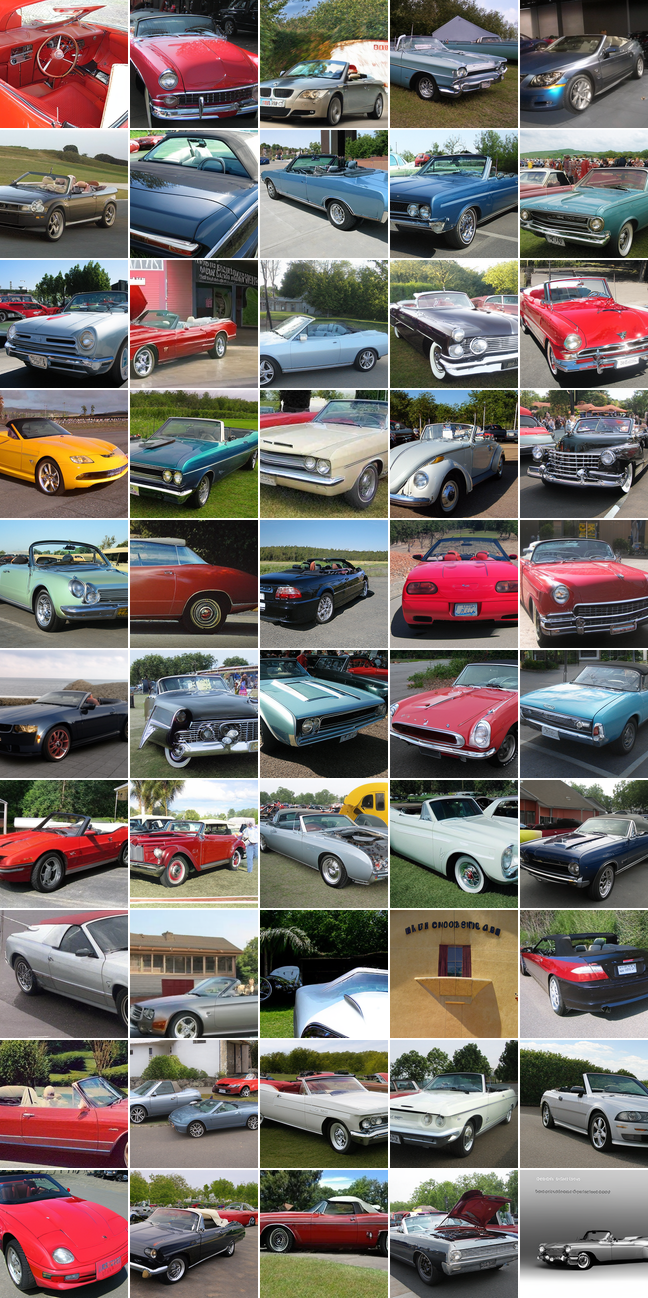}
        \caption{Class: ``convertible'' (511)}
        \label{fig:cockatoo}
    \end{subfigure}
    \caption{\textbf{Uncurated} 256×256 \textbf{VARd30 + refiner} samples. Classifier-free guidance scale = 2.5, topk = 500.}
    \label{fig:app_imagenet_uncurated_6}
\end{figure}

\Cref{fig:app_baseline_vs_refiner} shows additional results of VAR with and without our refiner across different base model scales.
\Cref{fig:app_imagenet_uncurated_1,fig:app_imagenet_uncurated_2,fig:app_imagenet_uncurated_3,fig:app_imagenet_uncurated_4,fig:app_imagenet_uncurated_5,fig:app_imagenet_uncurated_6} show uncurated ImageNet samples for various classes.

% \begin{figure}[t]
%     \centering
    
%     \begin{subfigure}{0.48\textwidth}
%         \centering
%         \includegraphics[width=\linewidth]{img/class22_refiner.png}
%         \caption{\textbf{Uncurated} 256×256 \textbf{VARd30 + refiner} samples. 
%         Classifier-free guidance scale = 2.5, topk = 500. 
%         Class: ``???'' (22)}
%         \label{fig:husky}
%     \end{subfigure}
%     \hfill
%     \begin{subfigure}{0.48\textwidth}
%         \centering
%         \includegraphics[width=\linewidth]{img/class96_refinerd30.png}
%         \caption{\textbf{Uncurated} 256×256 \textbf{VARd30 + refiner} samples. 
%         Classifier-free guidance scale = 2.5, topk = 500. 
%         Class: ``???'' (96)}
%         \label{fig:cockatoo}
%     \end{subfigure}

% \end{figure}

\end{document}